\documentclass[preprint,12pt,authoryear]{elsarticle}
\usepackage{amsmath, amssymb, amsfonts}
\usepackage{graphicx}
\usepackage{hyperref}
\usepackage{geometry}
\usepackage{graphicx}
\usepackage{subcaption}
\usepackage[title]{appendix}%
\usepackage{setspace}

\title{Supervised Learning for the $(s,S)$ Inventory Model with General Interarrival Demands and General Lead Times}
\date{}

\begin{document}

\begin{frontmatter}



\title{} 


\author{Eliran Sherzer} 

\affiliation{organization={Department of Industrial Engineering and Management,  Ariel University},
            addressline={65 Ramat HaGolan}, 
            city={Ariel},
            postcode={407000}, 
            country={Israel}}

            \author{Yonit Barron} 

\affiliation{organization={Department of Industrial Engineering and Management,  Ariel University},
            addressline={65 Ramat HaGolan}, 
            city={Ariel},
            postcode={407000}, 
            country={Israel}}

\begin{abstract}
The continuous-review $(s,S)$ inventory model is a cornerstone of stochastic inventory theory, yet its analysis becomes analytically intractable when dealing with non-Markovian systems. In such systems, evaluating long-run performance measures typically relies on costly simulation.

This paper proposes a supervised learning framework via a neural network model for approximating stationary performance measures of $(s,S)$ inventory systems with general distributions for the interarrival time between demands and lead times under lost sales. Simulations are first used to generate training labels, after which the neural network is trained. After training, the neural network provides almost instantaneous predictions of various metrics of the system, such as the stationary distribution of inventory levels, the expected cycle time, and the probability of lost sales. We find that using a small number of low-order moments of the distributions as input is sufficient to train the neural networks and to accurately capture the steady-state distribution. Extensive numerical experiments demonstrate high accuracy over a wide range of system parameters. As such, it effectively replaces repeated and costly simulation runs. Our framework is easily extendable to other inventory models, offering an efficient and fast alternative for analyzing complex stochastic systems. 
\end{abstract}




\end{frontmatter}

\section{Introduction}

In the twentieth century, supply chain management underwent significant changes due to fluctuating demand, fragmented distribution channels, short product life cycles, rapid changes in the economic environment, and increased competitiveness. 
Inventory management is a critical component of retail supply chains, directly affecting operational efficiency, customer satisfaction, and profitability. Traditional approaches to inventory optimization often rely on heuristic rules or mathematical models that make strong assumptions, which struggle to cope with real-world, stochastic modern retail environments~\cite{park2025inventory}. 

This paper proposes a novel framework that uses Machine Learning (ML) techniques to optimize inventory control decisions.  We assume an inventory system under general interarrival times between demand instances and general lead times. Each demand requires one unit, and any unsatisfied demand is lost (thus, we have non-negative inventory levels). It is well known that if interarrival times or lead times are exponentially distributed, deriving the cost of a specific policy becomes somewhat simple by using Markov renewal techniques and embedded processes (see, e.g., ~\cite{barron2020qmcd}). If, however, general distributions are introduced, finding optimal or near-optimal policies is analytically complex. Hence, an alternative method should be considered that captures the stochastic nature of the process both quickly and accurately.

The classical $(s,S)$ inventory policy is the cornerstone of base-stock inventory policies, applicable in numerous operational research and logistics scenarios. Under $(s,S)$, the inventory is refilled up to level $S$ whenever its on-hand inventory level reaches or falls below $s$. Prior research shows that assuming such a variable order size after a random lead time is particularly practical in numerous scenarios, e.g., in vendor-managed contracts, limited shelf space, and powerful retailers. Real-world examples include a gas supplier sending trucks to fill up the tanks of the customers, or a retailer filling food vending machines in public buildings (e.g., hospitals and academic institutions). An excellent comprehensive survey on continuous-review $(s,S)$ inventory policies is found in ~\cite{perera2023survey}. 

The model accounts for three different cost components. Holding costs capture the expense of carrying inventory and are assumed to increase linearly with the inventory level over time. Ordering costs are incurred whenever a replenishment order is issued. Lost-sale costs arise when demand occurs while no inventory is available.

The cost components give rise to three key stationary performance measures of the underlying inventory process. The first is the stationary inventory-level distribution, which determines the holding cost. The second is the stationary expected cycle time, defined as the duration between consecutive replenishment epochs, and governs the ordering cost. The third is the stationary probability of a lost sale, when the inventory level is zero, and demand arrives. Obtaining those stationary measures in an $(s,S)$ system is essential to evaluate the long-term operational performance, thus enabling comparison and optimization of the thresholds $(s,S)$ and cost analyses. We note that the classic analytic approach to derive these measures is limited to a stylized model that requires unrealistic assumptions, such as the Markovian assumption on the interarrival times or lead times. Currently, the only available method to analyze non-Markovian systems is via simulation, which suffers from long convergence times.

With technological advancements, the application of ML algorithms in this field has aroused increasing interest, offering substantial opportunities for process optimization and addressing the inherent challenges of inventory control. We present a supervised ML framework that approximates the steady-state probabilities almost instantaneously. Specifically, we implement neural networks (NNs) as our ML framework. The choice of NNs stems from their previous successes in various stochastic models with similar characteristics. For example, computing the steady-state probabilities of the queue length in a $GI/GI/1$ queue~\cite{sherzer23} is to some extent equivalent to calculating the steady-state distribution of the inventory level.

In both cases, the stochastic process is determined by two independent and non-negative distributions. In inventory models, the interarrival times between demands and the lead times; in queueing models, the interarrival times between customers and service times. However, in contrast to the $GI/GI/1$ queueing model, the $(s,S)$ model depends critically on two control parameters, which introduce several challenges. In particular, the parameter $S$ alters the practical support of the system, substantially complicating the training process. Furthermore, the $(s,S)$ inventory model introduces discontinuities and regime changes that are absent in other models (e.g., queueing models), where the state evolves freely without reflection or resetting.

Implementing NNs in inventory systems poses three key challenges. First, generating a sufficiently diverse set of input instances to the system, including interarrival times between demands and lead times. The trained data must adequately cover the space of non-negative distributions to ensure broad generalization. Second, producing accurate output labels is challenging because they arise from a complex and untraceable stochastic process. Third, representing continuous distributions in a discrete format that allows the use of NNs. Distributions must be converted into fixed-size tensor inputs, a process that increases the risk of information loss.

To address the first challenge, we use the Phase-Type (PH) family of distributions, which is dense in the set of all nonnegative distributions. When general distributions are appropriate, phase distributions can be taken into account in a natural way, as any nonnegative continuous probability distribution can be approximated with a phase-type distribution (Theorem 4.2, Chapter III of ~\cite{Asmussen2003}). This enables structured yet diverse sampling from the space of input distributions. We note that sampling valid PH distributions is itself non-trivial due to interdependent parameter constraints. To overcome this, we adopt the sampling approach introduced in~\cite{sherzer23}, which efficiently explores the PH space while maintaining distributional diversity. 

To address the second challenge, accurate output labels, we use a discrete-event simulation. Since simulation is computationally expensive, it is performed offline during the data generation phase. This allows for the allocation of significant resources without real-time constraints. Once trained, the NN model enables rapid inference, producing predictions in milliseconds instead of the long and expensive simulation time while generalizing to unseen instances beyond the training set.

To address the third challenge, the input representation, we encode each distribution using its first $n$ moments. These moments, either computed analytically or estimated empirically, are converted to logarithmic values to reduce numerical instability. The specific choice of $n$ is determined as a tunable hyperparameter. We note that choosing the number of moments $n$ plays a critical role in the accuracy of the model. Prior work~\cite{sherzer23,  SHERZER2024, SHERZER2025141, sherzer2025analyzinghomogenousheterogeneousmultiserver} indicates that, in stationary queueing settings, the first four to five moments generally suffice to capture stochastic behavior effectively. In Section~\ref{sec:mom_anal_res}, we perform a similar analysis and show that incorporating moments beyond the fifth does not lead to further improvements in the model's accuracy. Here, we quantify the accuracy achieved by adding the moments of the interarrival time between demands and the lead times, providing valuable insight into the impact of the $n^{\text{th}}$ moment on the overall process.

Although this paper discusses a specific inventory control model, NNs are not limited to a specific model setting, nor are the simulations that label the data. Thus, our approach can be straightforwardly extended to other models. Section~\ref{sec:discussion} discusses possible extensions.

The main contributions are as follows. 
(i) We develop a framework for labeling data, training, and making inferences for a stochastic non-Markovian inventory model using NN algorithms. 
(ii) The present study quantifies the impact of the $n^{th}$ moment of interarrtime between demands and lead times on the stationary inventory level. By that, we provide a unique opportunity to examine the role of these moments. (iii) We provide an open-access package that can be used for inference for the underlying model (see \url{https://github.com/eliransher/inventory_AI.git}). Given the pair $(s,S)$ and the first five moments, our model efficiently approximates the stationary inventory level, the average cycle time, and the probability of loss. In doing so, we provide decision makers with an effective and easy-to-use tool that derives the optimal thresholds and maximizes profitability. 

The remainder of this paper is organized as follows. In Section~\ref{sec:literture} we review the relevant literature on continuous-review inventory models under the $(s,S)$ control policy, phase-type (PH) representations, and recent learning-based approaches for steady-state analysis in queueing and inventory systems. Section~\ref{sec:Problem formulation} introduces the $(s,S)$ inventory model and the relevant costs. Section~\ref{sec:sol_overview} provides a high-level overview of the proposed solution framework, which is detailed in Section~\ref{sec:training_proc}. Section~\ref{sec:experiments} describes the experimental design, test-set segmentation, and performance metrics used. Section~\ref{sec:results} is dedicated to the contribution of higher-order moments to the prediction performance. Section~\ref{sec:Example} numerically demonstrates how the proposed approach derives the optimal thresholds. Finally, Section~\ref{sec:discussion} summarizes the main findings and implications, and suggests directions for future research. 

In the appendices, we provide additional numerical details, including statistical summaries of the test set and supplementary figures that support the main experimental results.

\section{Literature Review}\label{sec:literture}
We next review some relevant work from inventory management literature and the rapidly growing field of neural network-based control and optimization. 

Inventory management is a critical area of research in operations research and supply chain theory. Classic models that assume fixed demand and delivery times are analytically tractable and yield closed-form solutions for simple scenarios. In the literature, the $(s,S)$ control policy is one of the most widely used~\cite{perera2023survey}. The optimality of $(s,S)$-type policies has been investigated for various inventory models, including those with discrete and continuous-time review, different time horizons, discounted or average cost criteria, and backlogging or lost-sales.
Under the continuous-time review, the underlying demand process is typically linked to either varying rates~\cite{barron2021triple}, fluid processes~\cite{kawai2015fluid}, MAP process ~\cite{barron2017shortage,chakravarthy2021queuing,anbazhagan2022queueing,vinitha2022analysis, barron2024shortage}, Wiener processes ~\cite{You19,gong2022inventory,yamazaki2020optimal}, superposition of deterministic rates combined with a renewal process ~\cite{presman2006inventory,benkherouf2009optimality,bensoussan2025optimality}, and other variables with a general distribution ~\cite{gurler2008analysis,barron2020qmcd}. For a comprehensive survey, see~\cite{perera2023survey}. 

With increasing demand variability, frequent sales, shifting consumer preferences, and multi-tiered chains, assumptions about certain distributions may not accurately reflect reality. Practically, most inventory management problems have complex dynamics, and finding optimal policies has been a major challenge. Thus, more adaptive methods have been developed using heuristic optimization and simulation-based approaches ~\cite{barron2019state,barron2024shortage}. For bibliographic surveys on data-driven models in performance evaluation and inventory simulation, see~\cite{beebe2025computing, beebe2025sigmetrics}. The simulation approach is conceptually simple yet capable of modeling complex production-inventory systems. However, the slowness for large-scale problems is its main and most noticeable drawback. To overcome this drawback, the use of ML algorithms in inventory control possesses the potential to analyze vast amounts of data, predict trends, and automate complex decisions, leading to enhanced and more effective optimization of stock levels~\cite{gutierrez2024benefits}. For a comprehensive review of the literature using ML algorithms in inventory control management, see, e.g.,~\cite{gutierrez2024benefits, ganjare2024systematic, garg2025impact}.

In particular, NNs can be widely utilized to simulate complex inventory systems and provide quick approximate solutions, although they are not always optimal~\cite{zhang2021study}.  There is a growing body of literature on the use of ML for inventory management optimization and decision-making. For example, Cevallos-Torres \textit{et al}.~\cite{cevallos2022decision} use NNs to deliver the value of estimated sales of soft drinks. The paper only utilizes NNs for predicting the demand, but does not deal with the complex task of capturing the nature of the stochastic system. 

Other papers use ML methods to propose decision-support frameworks. De Paula Vidal \textit{et al.}~\cite{de2022decision} propose a decision support framework for inventory management that combines multicriteria decision-making and ML approaches; their framework was applied in a railway logistics operator. Gijsbrechts \textit{et al}.~\cite{Gijsbrechts2022} utilize deep reinforcement learning algorithms to solve large Markov Decision Processes (MDPs) with heuristics for lost sales, dual sourcing, and multi-echelon inventory models. In their study, ML algorithms develop a good approximation of the value or policy function of the underlying MDP. Bottcher \textit{et al}.~\cite{bottcher2023control} contribute to this line of research by introducing a neural network-based optimization to the dual-sourcing inventory model that incorporates information on inventory dynamics and its control into the design of the neural network. Wang and Hong~\cite{wang2023large} propose a recurrent neural network-inspired simulation approach to solve an optimization problem of a large-scale production system with a general BOM production structure, where demands and inventories are reviewed periodically. In~\cite{DEHAYBE2024433}, the authors used deep reinforcement learning to obtain near-optimal dynamic policies for instances with a rolling horizon. In ~\cite{Gijsbrechts2022,bottcher2023control,wang2023large,DEHAYBE2024433}, the ML method replaces the role of MDP but not the estimation of the inventory distribution, as done here. For approximating the statistical invariance measures, simulations are still being used, which slows down the process.

In contrast, 
Piore \textit{et al.} ~\cite{priore2019applying} use NNs to approximate stationary measures as done here. They employ an inductive learning algorithm to set the most appropriate replenishment policy over time by reacting to environmental changes. Unlike the papers above, they replace simulations by NNs in order to obtain the mean cycle time, the mean inventory level, and the probability of lost sales. The main difference between our models and~\cite{priore2019applying} is that our models are trained over a wide range of interarrival time and lead time distributions, while in~\cite{priore2019applying} the lead time is deterministic and the time between demands is normal.

In summary, the existing literature does not offer a reliable method for analyzing non-Markovian stochastic inventory models, and it relies heavily on slow simulations. This paper proposes a framework to address this gap.  

\section{Problem Formulation}\label{sec:Problem formulation}
Consider an infinite-horizon single-item inventory system under a continuous-review, lost-sales, and $(s,S)$ type control policy. We assume that each demand requires one unit of inventory, and any unsatisfied demand is lost. Under the $(s,S)$ policy, when the inventory level arrives at or below the reorder point $s$, an order is placed to bring the inventory up to level $S$, after a random lead time. The total cost includes ordering costs for each replenishment (a fixed component and a variable cost that is proportional to the size of the replenishment), holding costs, and costs of unsatisfied demand (loss costs). Our objective is to derive the optimal thresholds $S^*$ and $s^*$ that minimize the average total cost per time unit. We use the following notation:

\begin{itemize}
\item Let $D_j,j=1,2,...$ be the interarrival time between the $j^\textit{th}$ and the $(j-1)^\textit{th}$ consecutive demands (with $D_0=0$). We assume that $D_j,j=1,2,...$ are independent and identically distributed (i.i.d) random variables (r.v.s) having a general distribution. Denote by $D$ the generic form of $D_i$ and let $m_D^i = E[D^i]$ be its $i^{\text{th}}$ moment. 
\item Let $L_j,1=1,2,...$ be the lead time of the $j^\textit{th}$ replenishment. We assume that $L_j,1=1,2,...$ are i.i.d r.v.s with a general distribution. Denote by $L$ the generic form of $L_i$ and let $m_L^i =E[L^i]$ be its $i^{\text{th}}$ moment. 
\item Let $s$ and $S$ be the reorder point and the order-up-to level, respectively (so called thresholds).
\item Let $I = \{ I(t) \mid t \ge 0 \}$ be the on-hand inventory level under a steady-state condition; without loss of generality, we assume that $I(0) = S$.
\item Let $\mathbf{P}$ be an $S+1$ vector whose $j^{\text{th}}$ entry, $P_j$, is the steady-state probability of having $j$ on-hand units in the inventory, i.e., $P_j=P(I=j)$.
\item Let $\rho$ denote the ratio between the first moments of the lead time and the inter-arrival time, $\rho = \frac{m_L^1}{m_D^1}$.
\item Let $C$ denote the cycle time, i.e., the time between two successive replenishments.
\item Let $\pi^i_{\text{arrival}}$ be the probability of having $i$ units in the inventory as seen by an arriving demand.
\item Let $K_o$ be the fixed ordering cost per each replenishment and $c_r$ be the cost for each purchased unit. Note that the replenishment size is a random variable that depends on the inventory level at the time of replenishment.
\item Let $c_h$ be the holding cost per unit in the inventory per time unit.
\item Let $c_l$ be the loss cost of an unsatisfied unit.
\end{itemize}

The average total cost per time unit under the $(s,S)$ policy, $g(s,S)$, is given by: 

\begin{align}\label{eq:cost_func1}
g(s, S) =
c_h  E[I]
+
\frac{K_o+c_rE[S-I(C^-)]}{E[C]}
+
c_l  {\frac{1}{m_D^1}}  \pi^0_{\text{arrival}},
\end{align}
where $ E[I] =  \sum_{i=0}^{S} i  P_i$.

Here, $I(C^-)$ is the inventory level just before replenishment, and $(S-I(C^-))$ is the actual replenishment size. We further note that since each demand is exactly one unit, the average replenishment size per time unit, $\frac{E[S-I(C^-)]}{E[C]}$, is actually the average satisfied demand per unit of time. Thus, it can be expressed by ${\frac{1}{m_D^1}}(1-\pi^0_{\text{arrival}})$. Hence, Equation~\eqref{eq:cost_func1} becomes: 

\begin{align}\label{eq:cost_func}
g(s, S) =
c_h  E[I]
+
\frac{K_o}{E[C]}
+
\frac{c_r (1-\pi^0_{\text{arrival}})}{m_D^1}
+
\frac{c_l  \pi^0_{\text{arrival}}}{m_D^1},
\end{align}
 
Accordingly, Equation~\eqref{eq:cost_func} shows that the average cost is a function of the difference ($c_l$-$c_r$), which highlights the ratio between acquisition costs and loss costs. To derive the optimal $S^*$ and $s^*$ that minimize $g(s,S)$, we approximate $\mathbf{P}$, $E[C]$, and $\pi^0_{\text{arrival}}$ using the sets of moments $\{m_D^1, m_D^2, \ldots, m_D^{n_D}\}$ and  $\{m_L^1, m_L^2, \ldots, m_L^{n_L}\}$; here $n_D$ and $n_L$ represent the number of interarrival demand and lead time moments that we use, respectively. Once $\mathbf{P}$, $E[C]$, and $\pi^0_{\text{arrival}}$ are estimated, the cost function $g(s,S)$ can be approximated for optimization purposes. 

\section{Solution Framework}\label{sec:sol_overview}

\begin{figure}[h]
\centering
\includegraphics[scale=0.6]{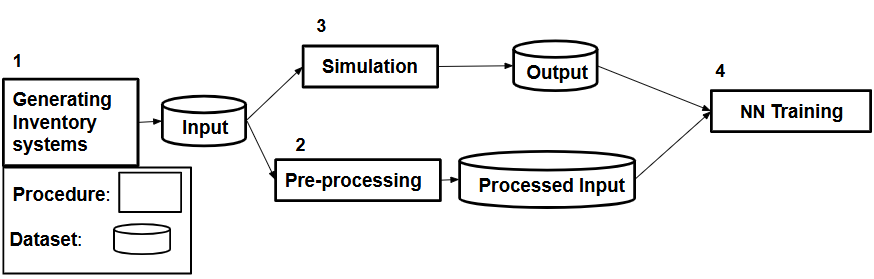}
\caption{Work-flow diagram of our learning procedure. }
\label{fig:diagram_overview}
\end{figure}

\begin{figure}[h]
\centering
\includegraphics[scale=0.55]{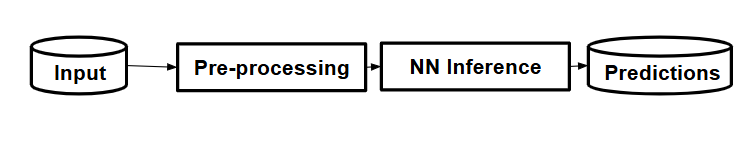}
\caption{Work-flow diagram for inference. }
\label{fig:diagram_inf}
\end{figure}

The core stages of our methodology are presented in Figure~\ref{fig:diagram_overview} and Figure~\ref{fig:diagram_inf}. Figure~\ref{fig:diagram_overview} illustrates the training pipeline, and Figure~\ref{fig:diagram_inf} outlines the inference process. We commence with Steps 1–3 from  Figure~\ref{fig:diagram_overview}, which correspond to the generation of training data, as detailed in Section~\ref{sec:data}. 

\textit{Step 1: Generating Inventory Systems} (Section~\ref{sec:input_generation}) constructs the underlying inventory model. Each instance includes: (i) the interarrival time distribution $\mathbf{D}$, (ii) the lead time distribution $\mathbf{L}$, and (iii) the reorder point $s$ and the replenishment level $S$.

\textit{Step 2: Pre-processing} computes $\{m_D^1, m_D^2, \ldots, m_D^{n_D}\}$ and  $\{m_L^1, m_L^2, \ldots, m_L^{n_L}\}$; these moments are then standardized (Section~\ref{sec:pre_process}). Note that this standardization applies only to training and is not used during simulation; the values of $n_D$ and $n_L$ are treated as tunable hyperparameters.\footnote{Hyperparameters are predefined model settings not learned during training, such as moment counts, network depth, or hidden layer sizes.}

\textit{Step 3: Simulation} computes statistical measures via discrete-event simulation (Section~\ref{sec:output_generation}). The output serves as the ground truth for training.

\textit{Step 4: Training} uses the processed inputs and simulation outputs to train the NN. Once trained, inference proceeds as presented in Figure~\ref{fig:diagram_inf}, where the network receives $s$, $S$, $\{m_D^1, \dots, m_D^{n_D}\}$ and $\{m_L^1, \dots, m_L^{n_L}\}$ as inputs. These inputs are pre-processed as in training, and the network outputs the estimated steady-state distribution. The network is subsequently evaluated under various settings using a test set described in Section~\ref{sec:datasets}.

\section{Training Process}\label{sec:training_proc}

In this section, we first describe the data generation process (Section~\ref{sec:data}). We then present the NN architecture (Section~\ref{sec:network}) and finally, the NN loss function (Section~\ref{sec:loss_func}).

\subsection{Data generating}\label{sec:data}

In this section, we construct the datasets for training, finetuning, and evaluation, as we follow the framework suggested in Section~\ref{sec:sol_overview}.

\subsubsection{Input generation}\label{sec:input_generation}

The main idea is to obtain a training set as diverse as possible. The input data consists of three parts. The first two parts include the distribution of the time between demands and the distribution of the lead time; both are non-negative and continuous. The third part is the value of $s$ and $S$.  These values are generated as follows: we first sample $S$ from $U(1,S_{max})$ and then $s$ is sampled from $U(0,S)$. Without the loss of generality, and for convenience purposes, we assume that $S_{max}$ is limited to 30; clearly, the suggested framework can be applied in case of higher values. 

The more challenging task is the first two parts. Here, we adopt the sampling method introduced in~\cite{sherzer23} to generate the input distributions. Specifically, we sample from the Phase-Type (PH) family, which is dense in the space of non-negative distributions. We generate PH distributions up to an order of magnitude of 500; this property allows us to construct a diverse and representative training set. As shown in~\cite{sherzer23}, this method offers a broad distributional coverage and compares favorably to alternative sampling approaches.

Another parameter to consider is $\rho$. The idea is to allow for both slow and fast lead times on average compared to the time between demands. For that, we scale all systems so that the average time between demands will be 1 and $\rho = m_L^1$. Then, we assume that the lead time rate is uniformly distributed between 0.1 and 10. 

\begin{figure}[ht]
    \centering
    \begin{subfigure}[b]{0.48\textwidth}
        \centering
\includegraphics[width=\textwidth]{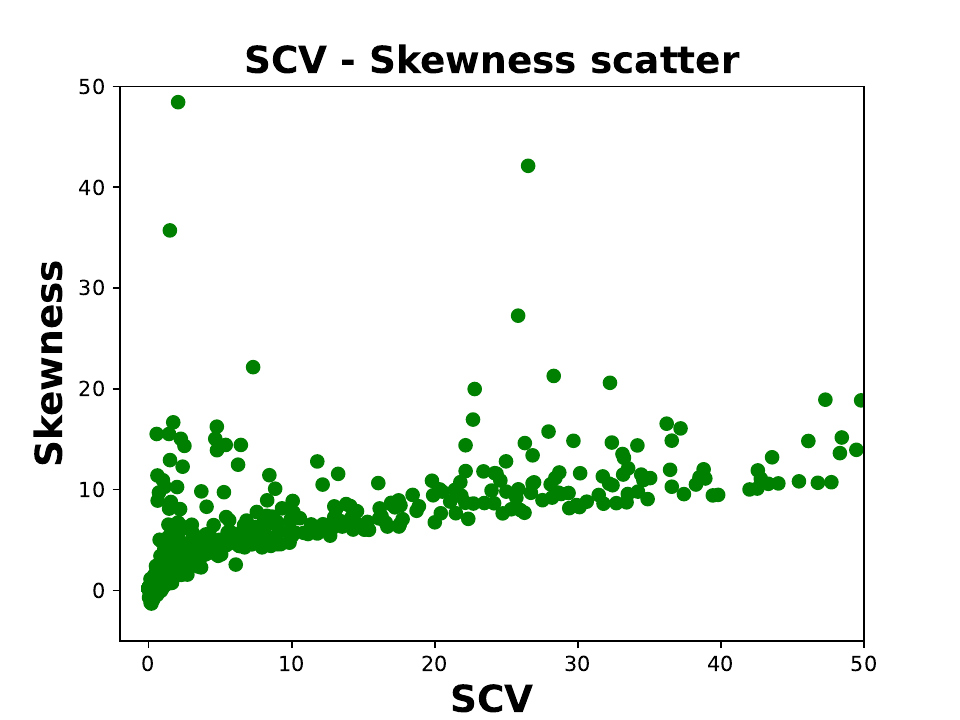}
        \caption{SCV-Skewness combination by different sampling techniques. }
        \label{fig:scv_skewness}
    \end{subfigure}
    \hfill
    \begin{subfigure}[b]{0.48\textwidth}
        \centering \includegraphics[width=\textwidth]{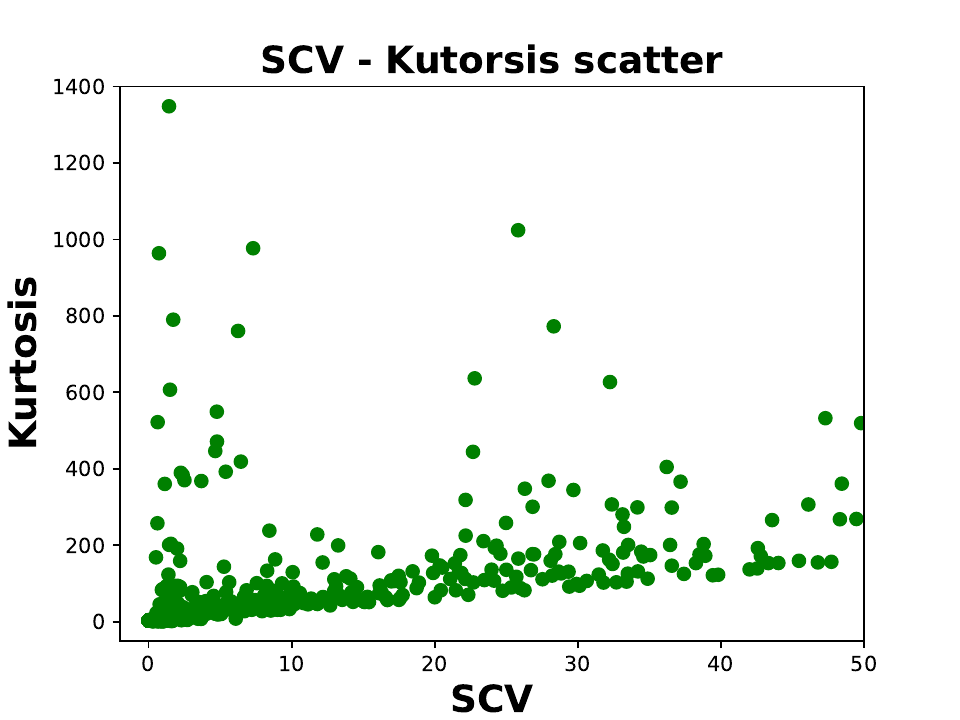}
        \caption{SCV-Kurtosios combination by different sampling techniques}
        \label{fig:scv_kurtosis}
    \end{subfigure}
    \caption{SCV, Skewness, and Kurtosis scatter.}
    \label{fig:scv_analysis}
\end{figure}

To illustrate the diversity of the data-generating process, we report the statistical measures of a set of 500 generated samples. The squared coefficient of variation (SCV) values span the range $[0.01, 50]$, skewness values range from $-1.3$ to $50$, and kurtosis values range from $1.08$ to $1350$. Figures~\ref{fig:scv_skewness} and~\ref{fig:scv_kurtosis} display the scatter plots of the generated PH distributions, showing SCV versus skewness and SCV versus kurtosis, respectively.

Although training relies solely on distributional moments rather than shape information, the generated PH distributions span a wide range of PDFs, including those with up to 6 distinct local maxima. More information regarding the shape of the generated distribution can be found in~\cite{sherzer23}.    
 
\subsubsection{Pre-processing Input }\label{sec:pre_process}

Once \textit{Step 1} is completed, these inputs undergo two pre-processing steps. The first step analytically computes the moments $\{m_D^1, m_D^2, \ldots, m_D^{n_D}\}$ and  $\{m_L^1, m_L^2, \ldots, m_L^{n_L}\}$. The second step applies a log transformation to the remaining moments in order to mitigate the wide range of moment values that may burden the learning process. This follows common practices in NN models to ensure that input values fall within a similar range. Notably, as the order of moments increases, their values grow exponentially, meaning that even after standardization, their range is not effectively constrained, and standardization adds computational complexity. Through experimentation, we find that the log transformation is both simpler to apply and more effective in reducing the variability of the moments. This also enjoys the empirical success from previuos studies~\cite{ sherzer23, SHERZER2024, SHERZER2025141}.

\subsubsection{Output generation}\label{sec:output_generation}

To label our data, we next perform simulations. For each instance, we simulate 180 million arriving demands. The simulations are written in Python via the package \textit{Simpy}. The simulation labels three statistical measures: the steady-state probability vector of the inventory level $\bold{P}$, the average cycle time $E[C]$, and the probability of lost sales $\pi^0_{arrival}$.   

Furthermore, to evaluate the accuracy of our simulation-based labeling, we conducted experiments on 500 systems. Each system was independently simulated 10 times, with each run processing 180 million arrivals. We then computed the lengths of the 95\% confidence intervals (CIs) for $E[I]$, $E[C]$, and $\pi^0_{arrival}$, which yielded 0.009, 0.26, and 0.0002, respectively; these results indicated high consistency across runs. (We note that each simulation required, on average, 0.44 hours on an Intel Xeon Gold 5115 processor with 128GB RAM.)

\subsubsection{Datasets: training, validation and test}\label{sec:datasets}

\noindent \textbf{Training and validation:} Training the NN requires two distinct datasets: a training set, used to optimize the model parameters, and a validation set, used for hyperparameter tuning. Both datasets are generated following the procedures outlined in Sections~\ref{sec:input_generation} and~\ref{sec:output_generation}. The training set consists of 2{,}500{,}000 instances, while the validation set contains 50{,}000 instances.

\noindent \textbf{Test Set:} To validate the model's accuracy, we construct a versatile test set, as described in Section~\ref{sec:datasets}. The test set is segmented by several key factors that influence the inventory dynamics. Specifically, we group instances based on the following three sets: (i) The SCV of $D$ and $L$ distributions, capturing the input variability. Here, each SCV classified by either low ($\text{SCV} < 1$) or high ($\text{SCV} \geq 1$). Such segmentation was also suggested in previous studies; see, e.g., ~\cite{sherzer23,SHERZER2024, You19}.

(ii) The values of $s$ and $S$ that determine the replenishment policy. The level $S$ is divided into two ranges: $S \in [1, 15]$ and $S \in (15, 30]$. For a given value of $S$, the point $s$ is classified into two intervals: $s \in [0, \lfloor S/2 \rfloor]$ and $s \in (\lfloor S/2 \rfloor, S]$. This multi-dimensional segmentation enables a comprehensive evaluation of the model's accuracy across a diverse set of operational regimes. We have five categories, each with two groups, for a total of 32 groups. 

(iii) The ratio $\rho$ that captures the relative speed of supply and demand. Here we distinguish between $\rho<1$ (i.e., a short time to replenishment) and $\rho>1$ (i.e., demand arrives faster). We note that such partitioning is analogous to underloaded and overloaded regimes of the traffic intensity parameter $\rho = \lambda / \mu$ in queueing theory (\cite{NAHMIAS1981141}).

In summary, we have five categories (SCV of $D$, SCV of $L$, $S$, $s$, and $\rho$), each with two ranges, for a total of 32 ranges.

Each of the 32 groups contains 1,500 samples, totaling 48,000 instances. Figures ~\ref{fig:rho_small} and \ref{fig:rho_large} in Appendix~\ref{append:scv_rho_testset1} present the values of $\rho$ for $\rho <1$ and $\rho >1$, respectively, and Figure~\ref{fig:scv_hist} presents histograms of SCV for $D$ and $L$ distributions. We see that the dataset is comprehensive due to its rich variability in SCV. Hence, the underlying sampling method yields greater diversity than alternative PH generation techniques (see also ~\cite{sherzer23}).

\subsection{Network Architecture}\label{sec:network}

\begin{figure}[h!]
\centering
\includegraphics[scale=0.75]{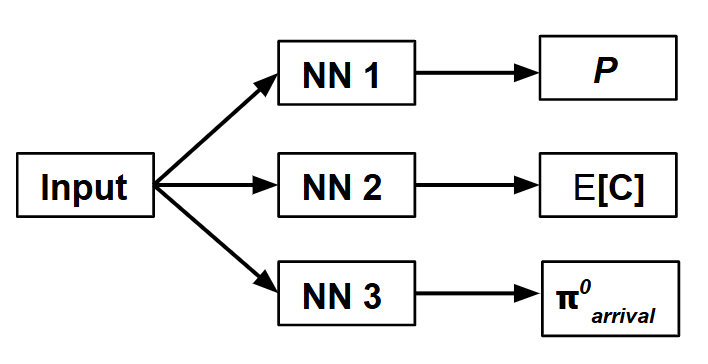}
\caption{Learning scheme. }
\label{fig:diagram_all_nns}
\end{figure}

Our next aim is to approximate $\textbf{P}$, $E[C]$ and $\pi^0_{arrival}$ using the moments of $D$, $L$, and the values $S$, and $s$. Denote by $NN$ 1, $NN$ 2 and $NN$ 3 the networks that approximate $\textbf{P}$, $E[C]$, and $\pi^0_{arrival}$, respectively (see Figure~\ref{fig:diagram_all_nns}).

We utilize a fully connected feedforward architecture to approximate the stationary distribution. We note that previous studies have also employed fully connected networks for similar tasks, e.g., the distribution of the queue length of different queueing systems (~\cite{sherzer23,SHERZER2025141}). However, they considered alternative architectures such as Convolutional Neural Networks (CNNs) and Recurrent Neural Networks (RNNs). In our model, these architectures were found to be unsuitable. In particular, CNNs are ineffective due to the low dimensionality of the input, and RNNs are more appropriate for time-dependent or transient analyses (~\cite{SHERZER2024}). Similarly, when referring to $E[C]$, and $\pi^0_{arrival}$.

\noindent For all NNs, all hidden layers use the Rectified Linear Unit (ReLU) activation function:
\[
\text{ReLU}(\text{input}) = \max\{\text{input}, 0\}.
\]

\noindent We next address the unique properties of each NN separately.

\noindent \textbf{NN 1:}  Here, we approximate the inventory steady-state distribution vector $\textbf{P}$. The output layer employs the following Softmax activation:
\[
\text{SOFTMAX}_i (a_1, \dots, a_n) = \frac{\exp(a_i)}{\sum_{j=1}^n \exp(a_j)}, \quad i = 1, \ldots, n,
\]

\noindent ensuring that the output values form a valid probability distribution summing to 1. However, since we approximate the distribution $\textbf{P}$, and the inventory level under the $(s,S)$ policy cannot exceed $S$, any probability mass assigned to levels above $S$ by the neural approximation must be removed. Yet, simply truncating the tail would lead to $\sum_{i=0}^{S} P_i< 1$. Thus, we require a \textit{support-constraint projection}. To preserve probabilistic consistency, we normalize the probabilities in the feasible domain $[0,S]$. This ensures that the corrected distribution represents a valid stationary probability mass function. To enforce feasibility, we define the corrected distribution $P^*_j$ as:
\[
P^*_j = 
\begin{cases}
\dfrac{P_j}{\displaystyle \sum_{i=0}^{S} P_i}, & 0 \le j \le S, \\[12pt]
0, & j > S.
\end{cases}
\]

\noindent This operation ensures that:
\[
\sum_{j=0}^{S} P^*_j = 1, 
\qquad P^*_j = 0 \ \text{for} \ j > S.
\]

\begin{figure}[h!]
\centering
\includegraphics[scale=0.5]{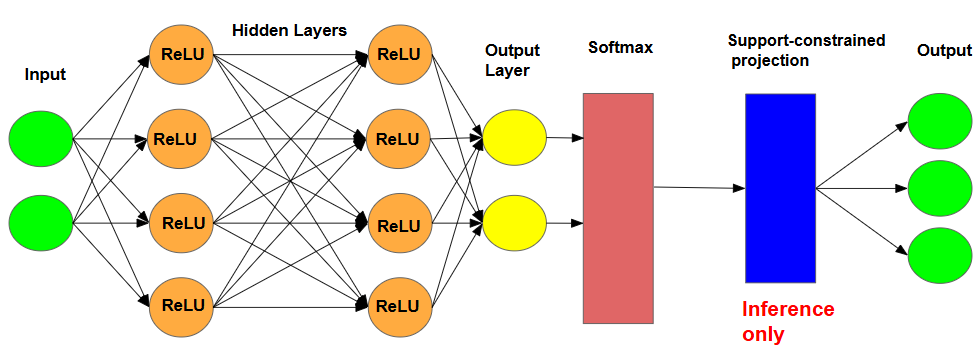}
\caption{\textbf{NN 1} architecture for approximating $\textbf{P}$. }
\label{fig:nn1_archi}
\end{figure}

\noindent We note that the support-constraint projection is only taking place for inference, and does not occur during training. The idea is to panelize the network when exceeding the current $S$. (see Section~\ref{sec:loss_func}). Figure~\ref{fig:nn1_archi} summarizes the architecture of \textbf{NN 1}.

\begin{figure}[h]
    \centering
    \begin{subfigure}[t]{0.48\textwidth}
        \centering
        \includegraphics[width=\textwidth]{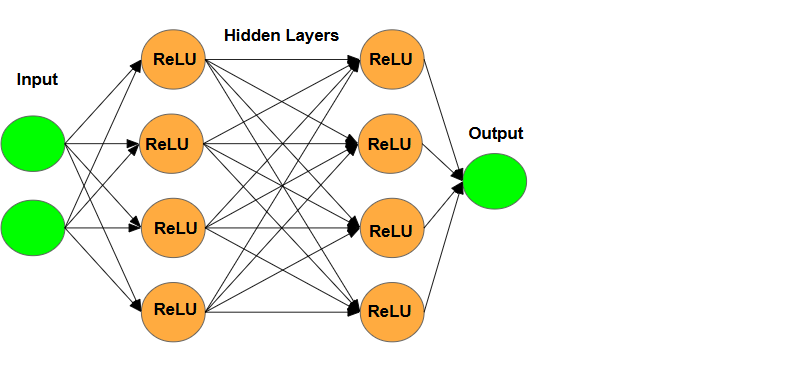}
        \caption{\textbf{NN 2} architecture for approximating $E[C]$ .}
        \label{fig:nn2}
    \end{subfigure}
    \hfill
    \begin{subfigure}[t]{0.48\textwidth}
        \centering
        \includegraphics[width=\textwidth]{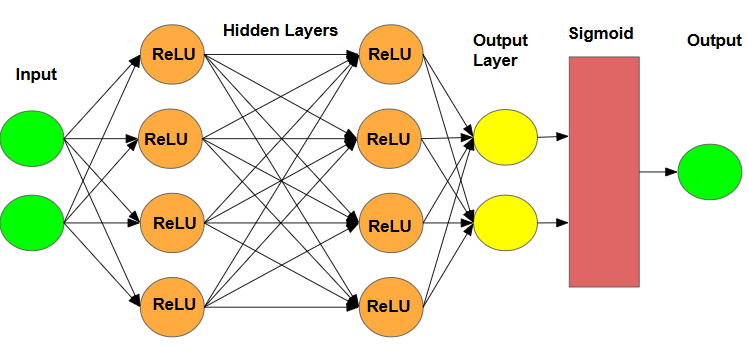}
        \caption{\textbf{NN 3} architecture for approximating $\pi^0_{arrival}$.}
        \label{fig:nn3}
    \end{subfigure}
    \caption{Architectures of \textbf{NN 2} and \textbf{NN 3}.}
    \label{fig:nn2nn3}
\end{figure}

\noindent \textbf{NN 2:} This network's goal is to approximate $E[C]$. Since it is a standard fully-connected NN, no additional activation function is required. Figure~\ref{fig:nn2} summarizes the \textbf{NN 2} architecture.

\noindent \textbf{NN 3:} This network's goal is to approximate $\pi^0_{arrival}$. The NN is implemented as a fully connected feedforward architecture with nonlinear activations in the hidden layers and a sigmoid activation function in the output layer. The sigmoid activation, defined as 
\[
\sigma(z) = \frac{1}{1 + e^{-z}},
\]
maps the unbounded latent output $z \in \mathbb{R}$ to the interval $(0,1)$, thereby 
ensuring that the network's prediction can be interpreted as a valid probability. 
In our context, the output $p = \sigma(z)$ represents the estimated probability 
of demand fulfillment under the $(s,S)$ policy, $1-\pi^0_{arrival}$. The use of the sigmoid corresponds 
to a logistic link function, providing a smooth, differentiable mapping between the 
latent score and the fulfillment probability while preserving probabilistic 
interpretability and numerical stability during training. The architecture is presented in Figure~\ref{fig:nn3}.

\subsection{Loss function}\label{sec:loss_func}

We present the loss function for all three NNs. The loss function derives a scalar measure of prediction error that quantifies the discrepancy between the network's predictions and the true target values. Minimizing the loss function during training guides the learning process to better fit the parameters to the actual model.

\noindent \textbf{NN 1:} The objective of \textbf{NN 1} is to approximate the distribution vector $\textbf{P}$. Let $\hat{\textbf{P}}$ denote the approximated vector of $\textbf{P}$. The network is trained using batch learning, where multiple data instances are processed together. Let $B$ represent the training batch size\footnote{The batch size is the number of training samples processed together in a single update of the model’s parameters; it balances the computational efficiency and the stability of gradient estimates.}, and $\mathbb{P}$ denotes the collection of $\textbf{P}$ vectors within a batch. The main challenge arises from the fact that each $\textbf{P}$ may correspond to a different $S$, leading to different vectors of varying lengths. To address this, each $\textbf{P}$ is padded with zeros to ensure that all vectors share the same dimension. Specifically, we define the dimension $l$ as the maximum value of $S$ in training plus one (since when the capacity is $S$, the inventory level can take values in $[0,...,S]$). In our training, we assume a maximum $S=30$ and, thus, $l = 31$.

Consequently, $\mathbb{P}$ is a $(B\times l)$ matrix, and $\hat{\mathbb{P}}$ denotes the corresponding matrix of approximated values generated by \textbf{NN 1}. We define the loss function as follows: 

\begin{align}
     Loss(\mathbb{P}, \hat{\mathbb{P}}) = \frac{1}{B} \sum_{i=1}^{B} \sum_{j=0}^{l} \big| \mathbb{P}_{i,j} - \hat{\mathbb{P}}_{i,j} \big|.
\end{align}

\noindent Note that several measures exist for the Loss function, e.g., squared differences and Kullback–Leibler divergence; however, they are less suitable for our setting. For example, using the squared differences formulation may seem appealing, but they have shown poor empirical performance. Because the loss function operates directly on probability values, squaring significantly reduces their impact on the gradient and the sensitivity of the training process. Accordingly, Kullback–Leibler divergence poses technical challenges when the distributions contain zero probability values (when inventory levels are higher than $S$). Although smoothing techniques can mitigate this problem, these adjustments can compromise numerical accuracy and introduce significant cumulative inaccuracies.

\vspace{0.5em}
\noindent \textbf{NN 2:} The objective of \textbf{NN 2} is to approximate $E[C]$. Due to its continuous feature, we use the mean squared error (MSE) measure for the loss function; this is the conventional choice for regression problems with continuous outputs.

\vspace{0.5em}
\noindent \textbf{NN 3:} The objective of \textbf{NN 3} is to approximate $\pi^0_{arrival}$. Similar to above, we employ MSE for the loss function. 
Although both \textbf{NN 1} and \textbf{NN 3} output probabilities, they are fundamentally different. \textbf{NN 3} outputs a single probability value with smooth and continuous deviations. In this case, MSE provides stable gradients and strongly penalizes large deviations, thereby accelerating convergence without facing significant numerical problems. In contrast, \textbf{NN 1} outputs a probability vector over discrete states. These probabilities, especially the tails, are usually small and sparse. As explained earlier, in this case, a quadratic operation will significantly reduce the sensitivity of the training. Appendix~\ref{append:train_proc} further provides technical details on the training process, such as the fitting process, learning-rate optimizer, etc.

\section{Experiments}\label{sec:experiments}

Once all NNs are trained, this section introduces the error matrices used for each NN; the results for the 32 segmentations (detailed in Section~\ref{sec:datasets}) are presented in Section~\ref{sec:results}.

\vspace{0.5em}
\noindent
\textbf{NN 1.} Two evaluation metrics are applied.

\vspace{0.5em}
1. Sum of Absolute Errors (SAE):
\begin{align*}
 \mathrm{SAE}(\mathbb{P}, \hat{\mathbb{P}}) 
    = \frac{1}{N} \sum_{i=1}^{N} \sum_{j=0}^{l} 
    \big| \mathbb{P}_{i,j} - \hat{\mathbb{P}}_{i,j} \big|.
\end{align*}

\noindent Here we denote by $\mathbb{P}$ and $\hat{\mathbb{P}}$ the true and predicted distributions in the (entire) test set, respectively (unlike Section~\ref{sec:loss_func}, where they only referred to the training batch). Accordingly, $N$ denotes the number of test instances, and $\mathbb{P}$ is an $(N\times l)$ matrix (recall that $l = max(S)+1=31$).

\vspace{0.5em}
2. Relative Error of the Mean (REM):
\begin{align*} \label{eq:REM}
\mathrm{REM} = 100 \cdot \frac{1}{N}\sum_{i=1}^N 
\frac{E[\mathbb{P}_i]-E[\hat{\mathbb{P}}_i]}{E[\mathbb{P}_i]},
\end{align*}
Accordingly, $E[\mathbb{P}_i] = \sum_{j=0}^{l} j\,\mathbb{P}_{i,j}$ and 
$E[\hat{\mathbb{P}}_i] = \sum_{j=0}^{l} j\,\hat{\mathbb{P}}_{i,j}$ are the true and predicted expected probabilities of having $i$ units in the inventory.

\vspace{0.5em}
\noindent
\textbf{NN 2.}  
To evaluate the expected cycle length, we use the metric Relative Error (RE) given by: 
\begin{align*}
\mathrm{RE}_{C} = 
   100 \cdot \frac{|E[C] - \widehat{E}[C]|}{E[C]}.    
\end{align*}

\noindent This metric normalizes deviations, allowing for a fair comparison between systems with different cycle time sizes.

\vspace{0.5em}
\noindent NN 3. To evaluate the probability of lost sales $\pi^0_{arrival}$, we use the metric Absolute Error (AE), given by:   
\[
    \mathrm{AE}_{\pi^0_{arrival}} = | \pi^0_{arrival}-\hat{\pi}^0_{arrival}|.
\]

AE avoids the instability of relative error metrics when the true probability p is very small, and may yield artificially large relative errors.

\section{Results}\label{sec:results}

This section presents the result of the moment analysis (Section~\ref{sec:mom_anal_res}), the accuracy (Section~\ref{sec:res_accuracy}), and runtime inference (Section~\ref{sec:runtimes}).

\subsection{Moment Analysis}\label{sec:mom_anal_res}

In this section, the influence of the number of moments on the prediction accuracy is studied. The objective is twofold. First, we aim to determine the smallest number of moments that will yield accurate and stable predictions across the three NNs. Second, and perhaps more importantly, we seek to gain a deeper understanding of the underlying stochastic process through its moment representation. By examining how the inclusion of successive moments affects model performance, we can assess which statistical features of the distributions are truly informative and which provide diminishing returns.

\begin{figure}[h!]
    \centering
    \begin{subfigure}[b]{0.32\textwidth}
        \centering
        \includegraphics[width=\textwidth]{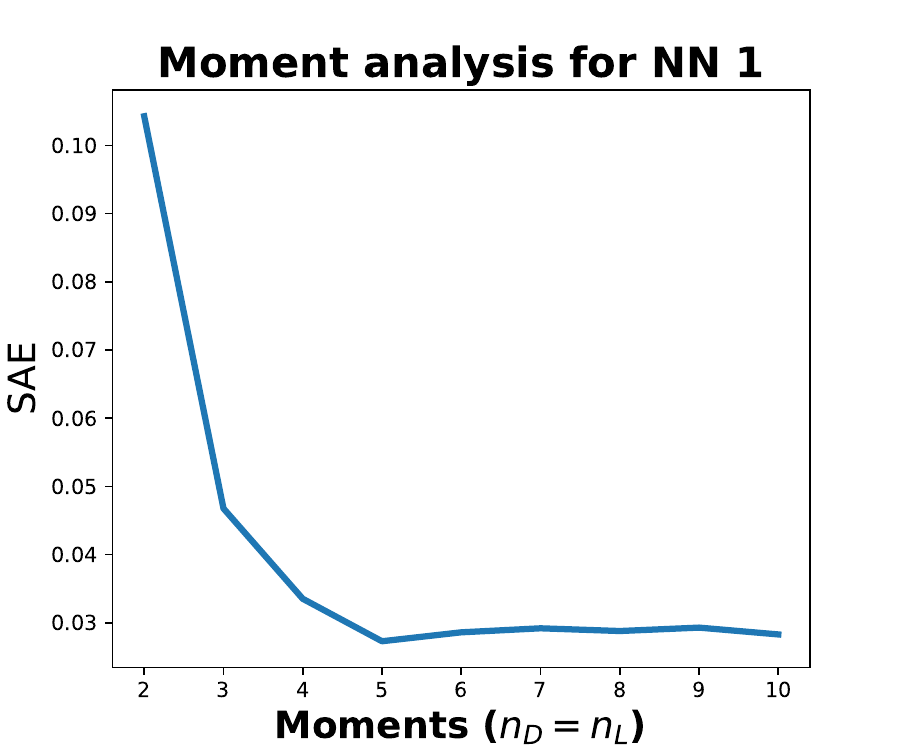}
        \caption{ \textbf{NN} 1.}
        \label{fig:mom_anal_nn1}
    \end{subfigure}
    \hfill
    \begin{subfigure}[b]{0.32\textwidth}
        \centering
    \includegraphics[width=\textwidth]{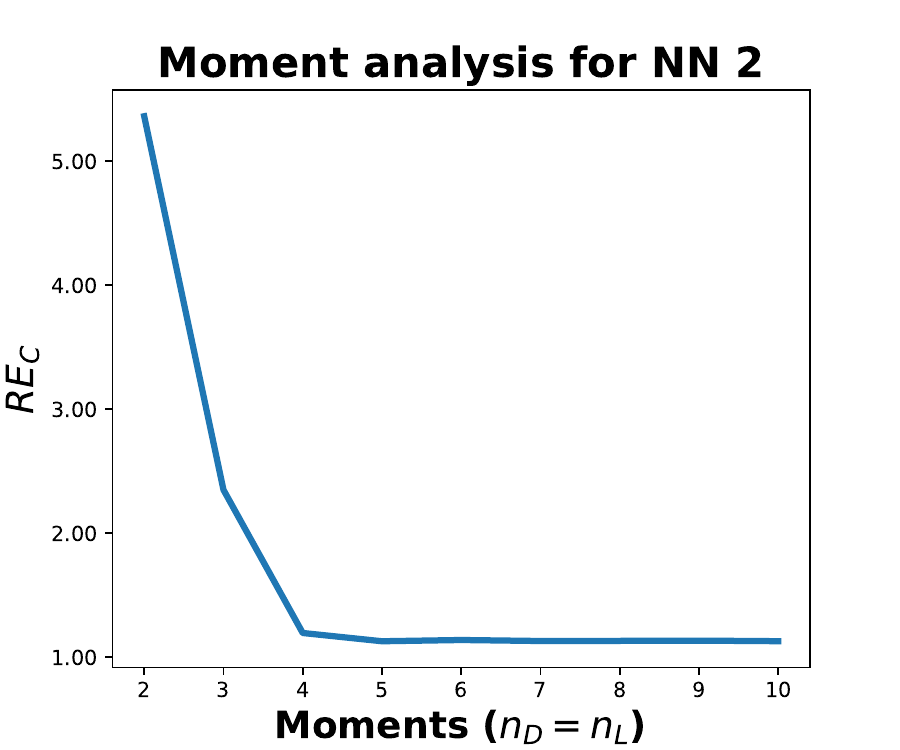}
        \caption{ \textbf{NN} 2.}
        \label{fig:mom_anal_nn2}
    \end{subfigure}
    \hfill
    \begin{subfigure}[b]{0.32\textwidth}
        \centering
        \includegraphics[width=\textwidth]{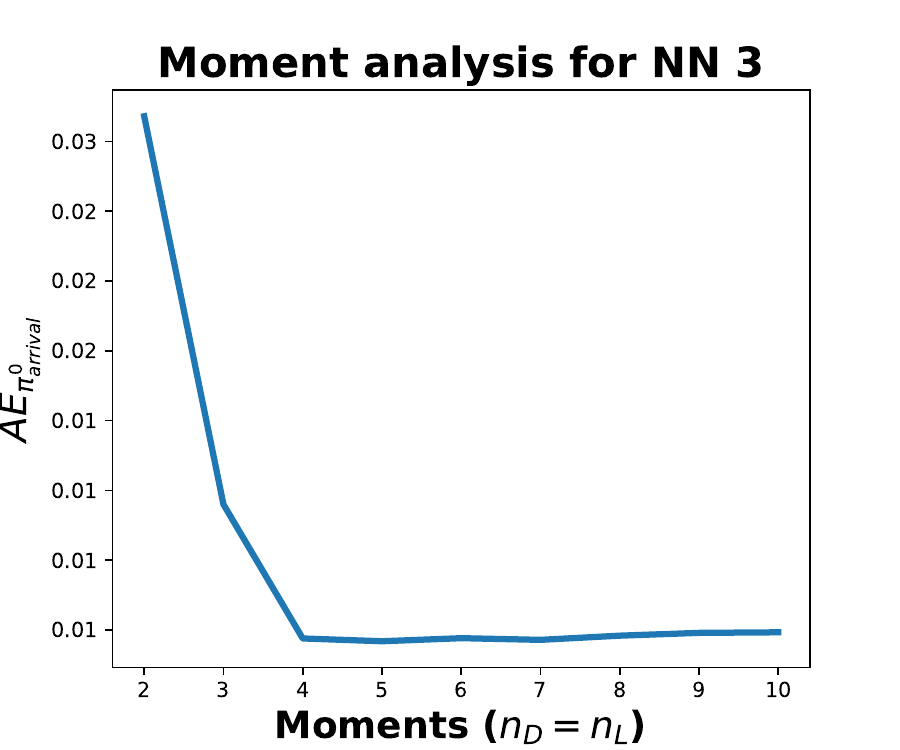}
        \caption{\textbf{NN} 3.}
        \label{fig:mom_anal_nn3}
    \end{subfigure}
    \caption{The error metrics as a function of $n_D=n_L$.}
    \label{fig:mom_anal_all}
\end{figure}

Figures~\ref{fig:mom_anal_nn1}--\ref{fig:mom_anal_nn3} present the error metrics (each in its relevant metric) as a function of the number of moments for \textbf{NN 1}, \textbf{NN 2}, and \textbf{NN 3}, respectively. The figures show that the accuracy improves steadily as the number of moments is increased up to five, at which point all three NNs achieve their best performance. Beyond this point, adding a sixth moment slightly degrades accuracy. This non-monotonic behavior suggests that higher-order moments introduce noise or instability that may obscure rather than improve the prediction signal.

Interestingly, this behavior reveals something about the stochastic nature of the system itself. While the first five moments (mean, variance, skewness, kurtosis, and the fifth moment) capture most of the process’s structural variability, the sixth and higher moments appear to encode negligible or redundant information. It is intriguing to observe how limited the contribution of higher-order moments is, implying that additional statistical details have little effect on the long-run behavior of the inventory process.
However, it is important to note that we are not claiming that the sixth moment and beyond are not informative or irrelevant. Higher-order moments may indeed capture fine-grained aspects of the distribution's shape that affect the system dynamics. 
Rather, our findings suggest that, within the current supervised learning framework, the NNs are unable to effectively leverage this additional information to improve predictive accuracy. This may stem from the nonlinear nature of the model or from the limited representational capacity of moment-based encoding.

We also note that these results are consistent with previous findings in queueing models, in which the first four to five moments were found to be sufficient to approximate the steady-state behavior of general stochastic systems; see ~\citep{sherzer23, SHERZER2025141}. 

In conclusion, our results indicate that the first five moments create a solid foundation that balances parsimony, accuracy, and practical learning.

\subsection{Accuracy}\label{sec:res_accuracy}
Table~\ref{tab:res_nn123} presents the result of \textbf{NN 1} (the columns SAE and REM), \textbf{NN 2} , and \textbf{NN 3}. We see that \textbf{NN 1} has generally high prediction accuracy in all regimes, with SAE $\in(0.01,0.03)$ and REM$<1.5\%$. The best accuracy is obtained in a low variability environment, where $\mathrm{SCV}_{\text{D}} \leq 5$, $\mathrm{SCV}_{\text{L}} \leq 5$, and when $\rho > 1$.  Larger errors are observed for $S > 15$ and $\mathrm{SCV}_{\text{D}} > 5$; since these scenarios create broader and flatter inventory distributions that are more challenging to approximate.  

The results of \textbf{NN 2} show good robustness overall, with relative errors below $1\%$ in most $\rho > 1$ scenarios. 
However, the accuracy decreases for high SCV values and when ($\rho \leq 1$ with $\mathrm{SCV}_{\text{D}} > 5$); here, the cycle time distribution becomes more skewed. The errors also increase for larger $S$ values, indicating that the mapping between thresholds and cycle times becomes more or less correlated as the replenishment quantities increase.

The results of \textbf{NN 3} are remarkably consistent across all regimes; 
AE$<0.02$ with minimal sensitivity to SCV, $\rho$, and the thresholds.  Slightly higher absolute errors occur in cases where both demand and lead time variability are large ($\mathrm{SCV}_{\text{D}}, \mathrm{SCV}_{\text{L}} > 5$), consistent with the broader uncertainty in system performance.

\vspace{0.5em}

Overall, the three networks demonstrate a strong fit over a wide range of model parameters. The networks are highly accurate for moderate to low variability and balanced environments; high variability and high $S$ values yield greater challenges in approximation, especially for predicting the stationary distribution and cycle time.

\begin{table}[!htp]\centering
\singlespacing
\caption{Accuracy results of \textbf{NN 1}, \textbf{NN 2} and \textbf{NN 3}}\label{tab:res_nn123}
\begin{tabular}{|l|r|r|r|r|r|r|r|r|r|r|} \hline
&\multicolumn{2}{|c|}{SCV } & & & &\multicolumn{2}{|c|}{\textbf{NN 1}} &\textbf{NN 2} &\textbf{NN 3} \\ \hline
Test Set &D &L &$\rho$ &S &s &SAE &REM &${RE}_c$ &$ {AE}_{\pi^0_{arrival}}$ 
\\ \hline
1 &$<=5$ &$<=5$ &$<=1$ &$<=15$ &small &0.02 &1.48 &0.72 &0.006 \\
2 &$<=5$ &$<=5$ &$<=1$ &$<=15$ &large &0.02 &0.63 &1.62 &0.006 \\
3 &$<=5$ &$<=5$ &$<=1$ &$>15$ &small &0.01 &0.67 &0.36 &0.004 \\
4 &$<=5$ &$<=5$ &$<=1$ &$>15$ &large &0.03 &0.88 &1.88 &0.003 \\
5 &$<=5$ &$<=5$ &$>1$ &$<=15$ &small &0.01 &0.94 &0.45 &0.006 \\
6 &$<=5$ &$<=5$ &$>1$ &$<=15$ &large &0.02 &0.79 &0.57 &0.007 \\
7 &$<=5$ &$<=5$ &$>1$ &$>15$ &small &0.01 &0.73 &0.23 &0.004 \\
8 &$<=5$ &$<=5$ &$>1$ &$>15$ &large &0.03 &0.70 &0.44 &0.006 \\
9 &$<=5$ &$>5$ &$<=1$ &$<=15$ &small &0.01 &1.38 &0.81 &0.004 \\
10 &$<=5$ &$>5$ &$<=1$ &$<=15$ &large &0.02 &0.81 &1.6 &0.005 \\
11 &$<=5$ &$>5$ &$<=1$ &$>15$ &small &0.01 &0.66 &0.31 &0.003 \\
12 &$<=5$ &$>5$ &$<=1$ &$>15$ &large &0.03 &0.97 &1.14 &0.003 \\
13 &$<=5$ &$>5$ &$>1$ &$<=15$ &small &0.02 &0.56 &0.46 &0.008 \\
14 &$<=5$ &$>5$ &$>1$ &$<=15$ &large &0.03 &0.77 &0.73 &0.010 \\
15 &$<=5$ &$>5$ &$>1$ &$>15$ &small &0.02 &0.78 &0.31 &0.004 \\
16 &$<=5$ &$>5$ &$>1$ &$>15$ &large &0.04 &0.74 &0.6 &0.007 \\
17 &$>5$ &$<=5$ &$<=1$ &$<=15$ &small &0.03 &0.50 &1.48 &0.012 \\
18 &$>5$ &$<=5$ &$<=1$ &$<=15$ &large &0.03 &0.91 &3.07 &0.013 \\
19 &$>5$ &$<=5$ &$<=1$ &$>15$ &small &0.02 &0.93 &0.87 &0.006 \\
20 &$>5$ &$<=5$ &$<=1$ &$>15$ &large &0.04 &0.73 &2.77 &0.003 \\
21 &$>5$ &$<=5$ &$>1$ &$<=15$ &small &0.01 &0.55 &0.83 &0.009 \\
22 &$>5$ &$<=5$ &$>1$ &$<=15$ &large &0.02 &0.76 &1.17 &0.009 \\
23 &$>5$ &$<=5$ &$>1$ &$>15$ &small &0.01 &0.80 &0.47 &0.005 \\
24 &$>5$ &$<=5$ &$>1$ &$>15$ &large &0.02 &0.78 &0.93 &0.006 \\
25 &$>5$ &$>5$ &$<=1$ &$<=15$ &small &0.02 &0.94 &1.81 &0.017 \\
26 &$>5$ &$>5$ &$<=1$ &$<=15$ &large &0.03 &1.03 &3.26 &0.006 \\
27 &$>5$ &$>5$ &$<=1$ &$>15$ &small &0.02 &0.69 &0.67 &0.003 \\
28 &$>5$ &$>5$ &$<=1$ &$>15$ &large &0.04 &0.73 &2.22 &0.003 \\
29 &$>5$ &$>5$ &$>1$ &$<=15$ &small &0.02 &0.85 &1.29 &0.009 \\
30 &$>5$ &$>5$ &$>1$ &$<=15$ &large &0.03 &0.78 &1.55 &0.012 \\
31 &$>5$ &$>5$ &$>1$ &$>15$ &small &0.02 &0.75 &0.77 &0.006 \\
32 &$>5$ &$>5$ &$>1$ &$>15$ &large &0.03 &0.79 &1.42 &0.008 \\ \hline
\end{tabular}
\end{table}

\begin{figure}[h!]
\centering
\includegraphics[scale=0.5]{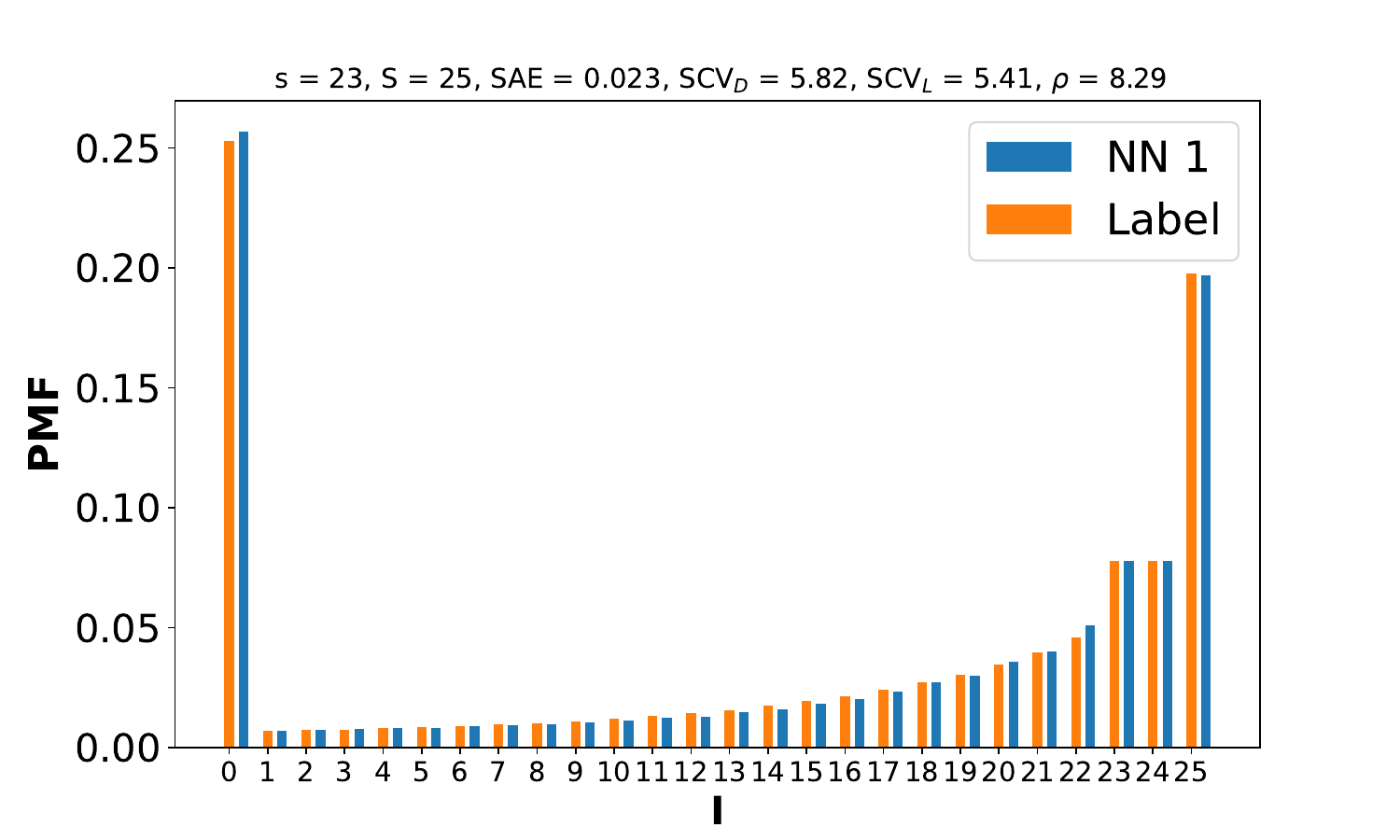}
\caption{\textbf{NN 1} results vs. simulation results in the exmaple. }
\label{fig:nn1_example}
\end{figure}

\begin{figure}[ht]
    \centering
    \begin{subfigure}[t]{0.48\textwidth}
        \centering
        \includegraphics[width=\textwidth]{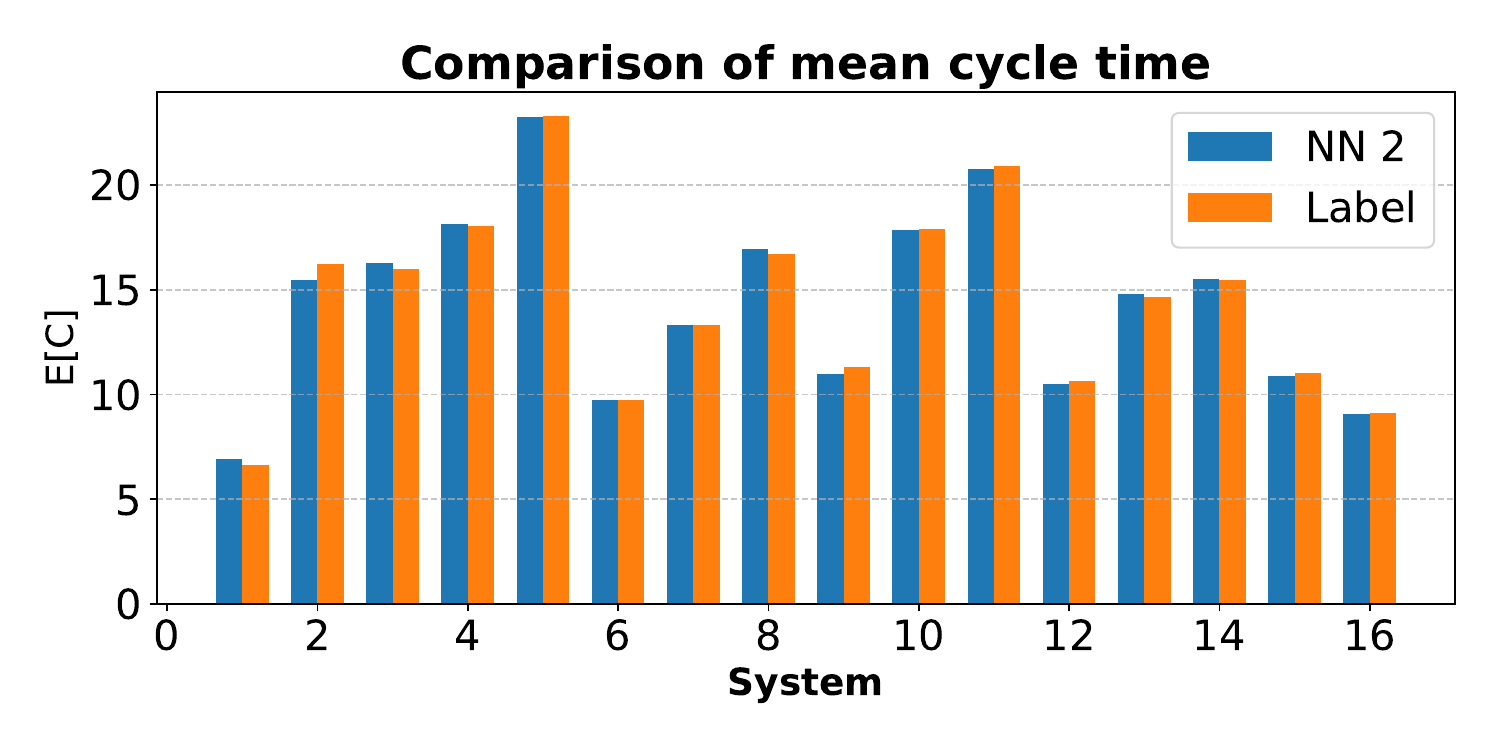}
        \caption{\textbf{NN 2} vs. simulation.}
        \label{fig:nn2_exmaple}
    \end{subfigure}
    \hfill
    \begin{subfigure}[t]{0.48\textwidth}
        \centering
        \includegraphics[width=\textwidth]{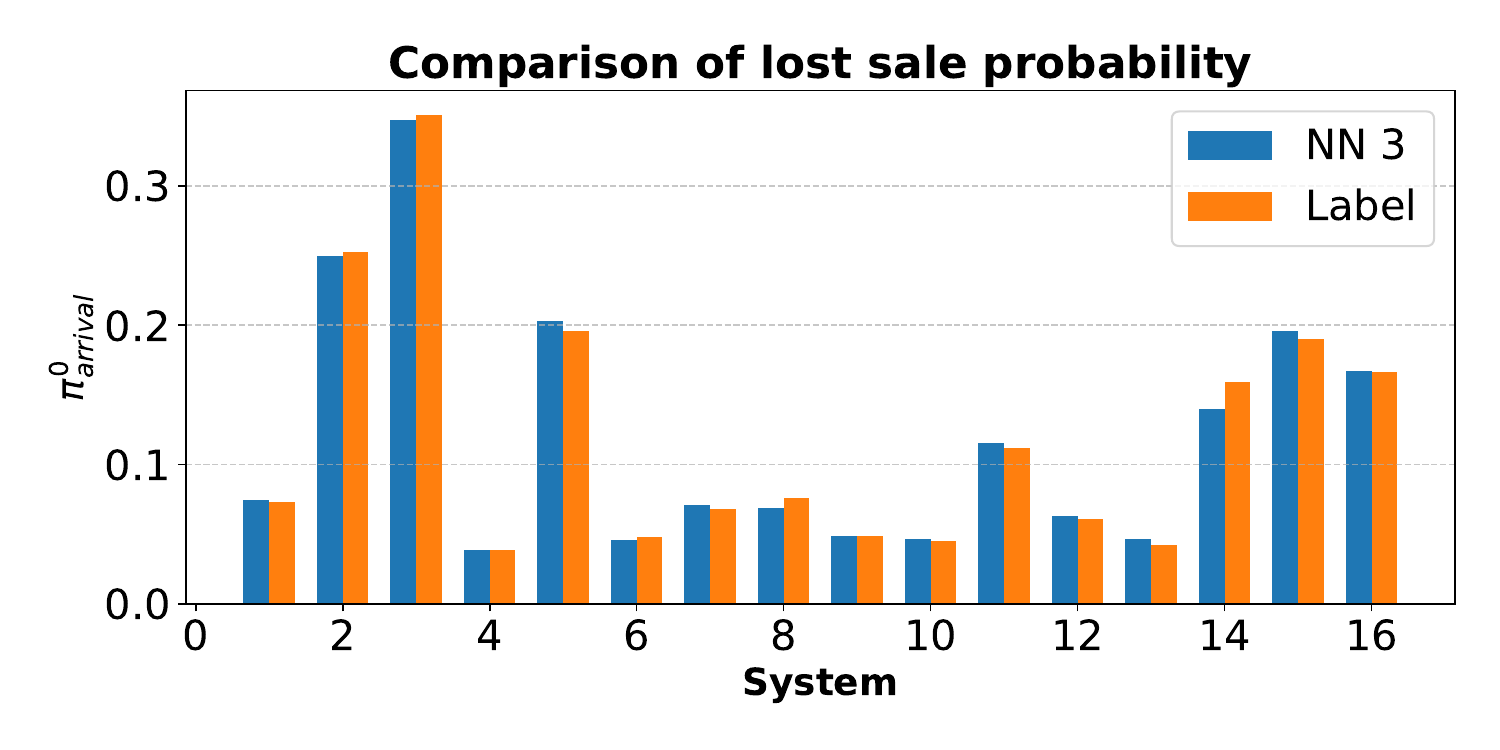}
        \caption{\textbf{NN 3} vs. simulation.}
        \label{fig:nn3_exmaple}
    \end{subfigure}
    \caption{\textbf{NN 2}, \textbf{NN 3} results vs. simulation results for the 16 different systems.}
    \label{fig:nn2nn3_examples}
\end{figure}

To illustrate, Figures~\ref{fig:nn1_example},~\ref{fig:nn2_exmaple}, and~\ref{fig:nn3_exmaple} visually present a numerical example of the results using \textbf{NN 1}, \textbf{NN 2}, and \textbf{NN 3}, respectively, compared to simulation results. Figure~\ref{fig:nn1_example} compares the predicted distribution of the inventory level; we see that the distributions are nearly identical (here, SAE$=0.023$). Figures~\ref{fig:nn2_exmaple} and~\ref{fig:nn3_exmaple} compare $E[C]$ and $\pi^0_{arrival}$, respectively, in 16 different systems, with the horizontal axis representing the output values predicted by the respective NN. We see that the prediction errors in both cases are minimal.

\subsection{Inference runtime}\label{sec:runtimes}

Table~\ref{tab:runtimes} presents the inference time for each NN. The runtimes representing the inference time for each NN are tested over a personal PC, with a processor Intel(R) Core(TM) i7-14650HX   2.20 GHz, 32 GB. One advantage of implementing deep learning networks is that they allow inference to be performed in parallel without increasing runtime. Therefore, we examine 1000 instances in parallel at a time, demonstrating the speed advantage over simulation. Table~\ref{tab:runtimes} presents the inference time for each NN; we see that for the 1000 instances, all NNs take only a fraction of a second.

\begin{table}[!htp]\centering
\caption{Inference runtimes}\label{tab:runtimes}
\begin{tabular}{|l|r|r|r|} \hline
\textbf{Network} &\textbf{ Instances} &\textbf{Runtimes} \\ \hline
NN 1 &1000 &0.015 \\
NN 2 &1000 &0.003 \\
NN 3 &1000 &0.027 \\ \hline
\end{tabular}
\end{table}

\section{Operational Example}\label{sec:Example}

In this section, we present a numerical example demonstrating how implementing the NN approach can efficiently obtain the optimal thresholds $(s^*,S^*)$ within a fraction of a second, as opposed to using simulations. 

Recall that the total cost $g(s,S)$ is a function of the costs $K_o$, $c_r$, $c_h$, and $c_l$ (see Equation~\eqref{eq:cost_func}). Let $c_h = 4$, $K_o = 100$, $c_r = 100$, and $c_l = 10{,}000$. Because our approach yields the entire inventory level distribution, it is relatively easy to add constraints, such as service level requirements. In this example, we assumed that the probability of having at least five units in stock would exceed $0.995$; i.e., $P(I \geq 5) \geq  0.995.$

\begin{figure}[h!]
\centering
\includegraphics[scale=0.5]{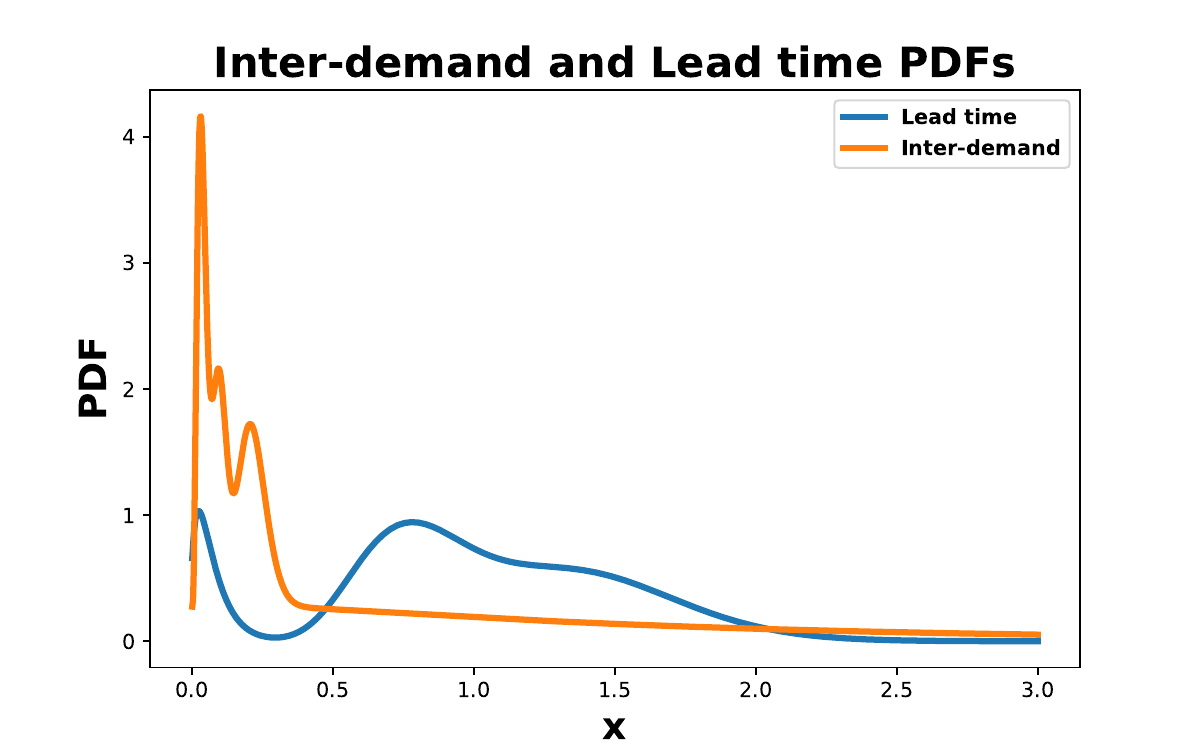}
\caption{D and L distributions. }
\label{fig:PDFs_numeric}
\end{figure}

Both $D$ and $L$ distributions are assumed to be PH distributions. Their statistical metrics, i.e., the mean, SCV, skewness, and kurtosis, are presented in Table~\ref{tab:numeric_props}. The corresponding probability density functions (PDFs) are shown in Figure~\ref{fig:PDFs_numeric}. 
As is evident, both distributions exhibit complex, non-exponential behavior, highlighting the flexibility of PH modeling.

\begin{table}[!htp]\centering
\caption{Statistical metrics of \textit{D} and \textit{L}}\label{tab:numeric_props}
\begin{tabular}{|l|r|r|r|r|r|}\hline 
&Mean &SCV &Skewness &Kurtosis \\ \hline
D &1 &2.35 &2.94 &18.85 \\
L &0.2 &0.26 &0.05 &2.36 \\ \hline
\end{tabular}
\end{table}

\begin{figure}
    \centering
    \begin{subfigure}[b]{0.48\textwidth}
        \centering
\includegraphics[width=\textwidth]{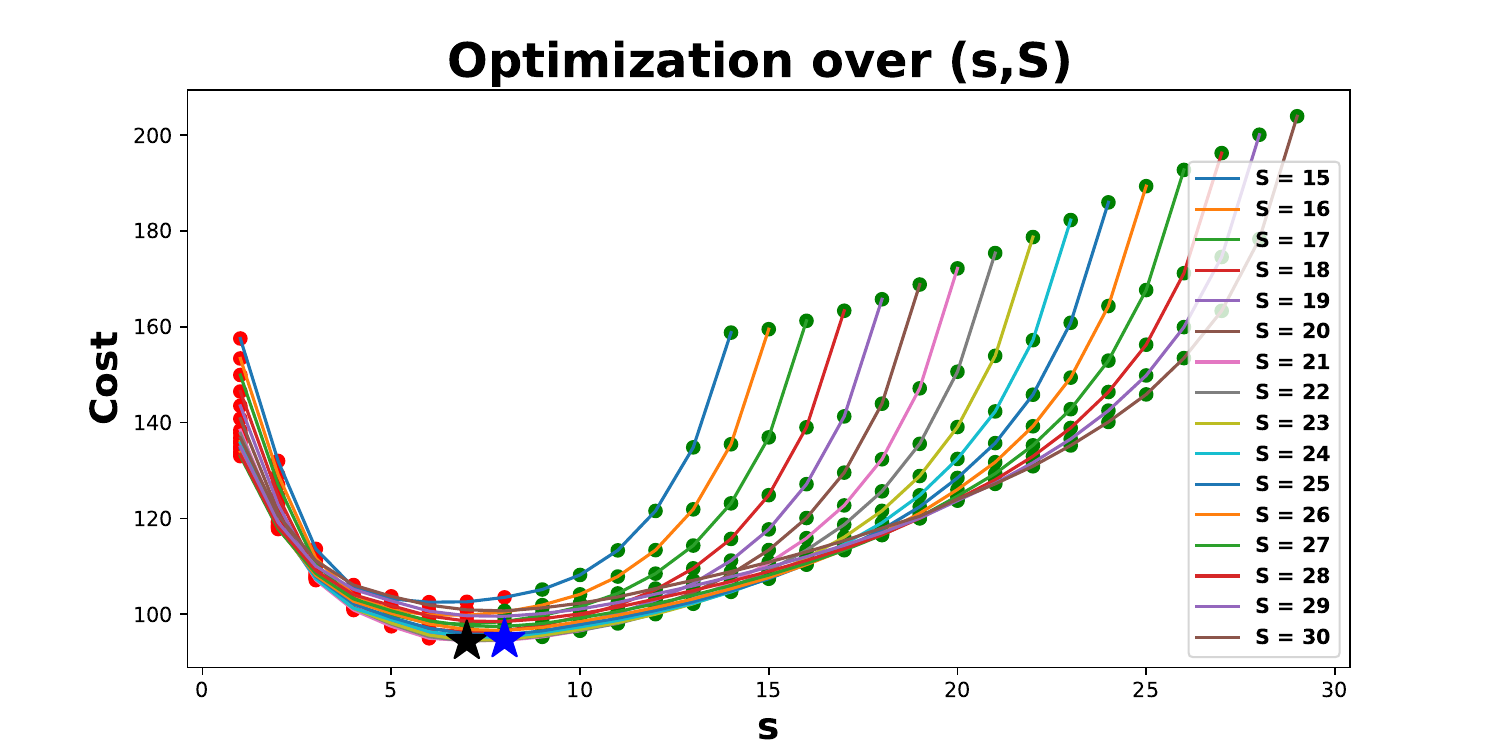}
        \caption{The total cost for different $(s,S)$. }
        \label{fig:zoomout}
    \end{subfigure}
    \hfill
    \begin{subfigure}[b]{0.48\textwidth}
        \centering \includegraphics[width=\textwidth]{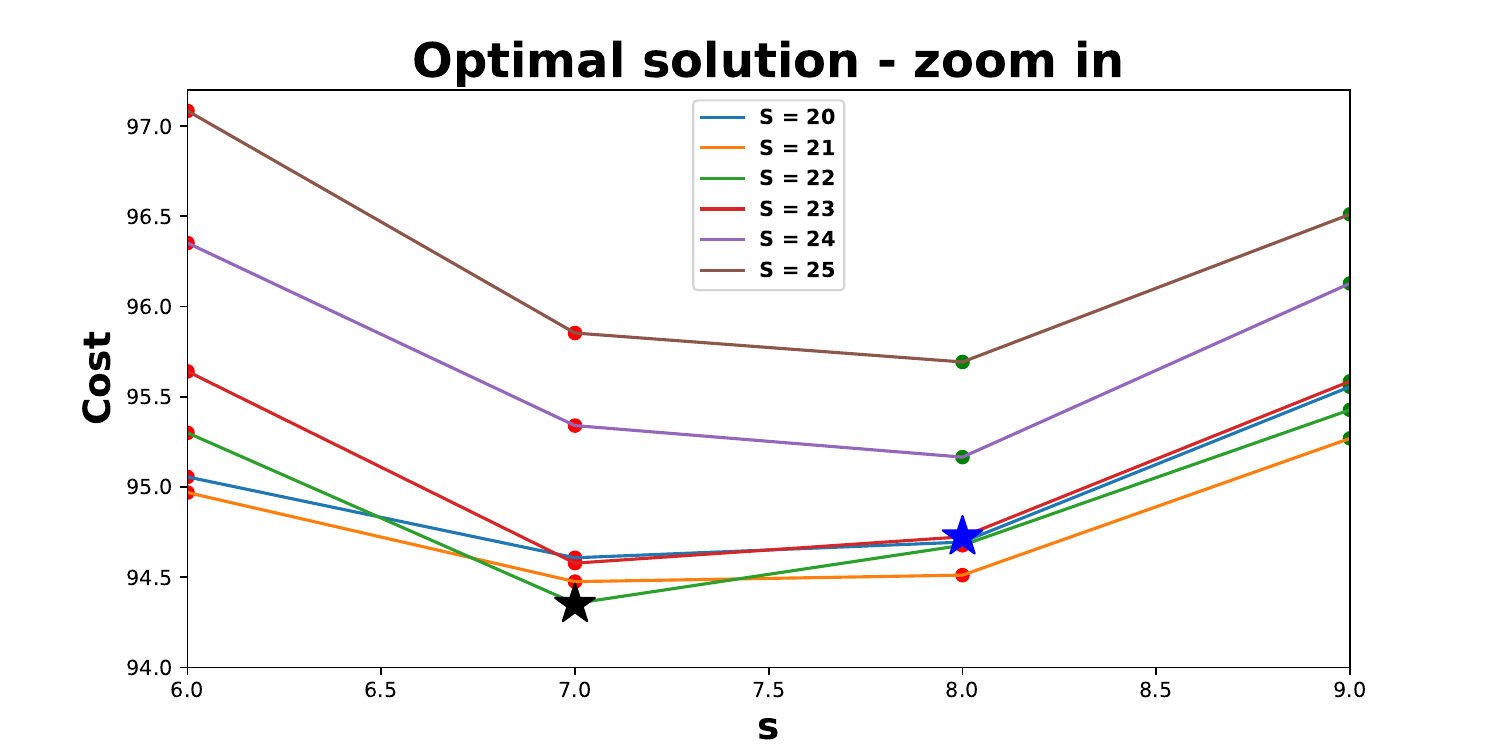}
        \caption{The optimal solution environment.}\label{fig:zoomin}
    \end{subfigure}
    \caption{The total cost function. The black (blue) star indicates the unconstrained (constrained) optimal thresholds. }
    \label{fig:scv_analysis}
\end{figure}

The results are summarized in Figures~\ref{fig:zoomout} and~\ref{fig:zoomin}. 
Figure~\ref{fig:zoomout} curves $g(s,S)$ as a function of $s=1,...,S-1$ for $S=15,...,30$ (results for $S < 15$ were omitted since $S^*$ is found to be outside this range). Figure~\ref{fig:zoomin} focuses on the optimal solution environment; all feasible pairs of $(s,S)$ are marked with green dots, and unfeasible pairs are marked with red. The black (blue) star indicates the unconstrained (constrained) $s^*$ and $S^*$. We see that the optimal policy is $(s^*,S^*)=(7,22)$ (see the black star); however,  this policy violates the service-level constraint. Under this constraint, the best feasible solution is $(s^*,S^*)=(8,23)$ (see the blue star). 
(We further note that for $S \leq 4$, the pairs $(s,S)$ are known \emph{a priori} to be unfeasible, due to the constraint of holding more than five units. However, for the sake of completeness, these cases were also evaluated.

The total computational runtime of evaluating 465 pairs $(s,S)$ required less than $0.004$ seconds, clearly demonstrating the practical efficiency and scalability of the method. For comparison, we used simulations and removed all pairs that were not feasible \emph{a priori}, leaving 459 feasible pairs. The optimal solution remained $(s^*,S^*) =( 8, 23)$, however, it required about $0.42$ hours of computation time.

\section{Discussion and Conclusions}\label{sec:discussion}

This paper proposed a supervised learning tool for approximating the stationary performance measures of a continuous-review $(s,S)$ inventory system with general interarrival times between demands and general lead times. By combining large-scale simulation-based labeling with NN approximations, we enable a fast and accurate prediction of the stationary distribution of the inventory level, the expected cycle time, and the probability of lost sales. These quantities are computationally expensive to obtain under non-Markovian assumptions.

A key advantage of the proposed framework is its model-agnostic nature. Since the expected targets are generated via simulation rather than analytical derivations, our approach is not limited to Markovian or closed-form models. Extending the framework to alternative replenishment policies, cost structures, periodic-review models, or operational constraints requires only the ability to simulate the system and generate labeled data, without modifying the learning architecture.

Furthermore, the proposed NNs are well-suited to capture complex nonlinear relationships between system parameters and long-run performance metrics, even in environments where the underlying stochastic dynamics exhibit regime changes or discontinuities. As a result, once trained, the network provides almost instantaneous predictions of steady-state characteristics over large parameter spaces, thereby effectively replacing costly repeated simulation runs. Practically, this framework allows decision makers to evaluate and compare inventory policies in real time in various stochastic environments, where traditional simulation-based approaches are impossible to implement. Importantly, we do not claim that the method proposed here aims to replace simulation as a modeling tool, but rather to leverage offline simulation in order to build fast and useful systems for inference and optimization.

In conclusion, this paper demonstrates that simulation-driven supervised learning offers a viable and scalable alternative for analyzing complex and non-Markovian inventory systems. By decoupling performance evaluation from analytical tractability, the proposed framework opens the door to systematic and efficient analysis of a broad class of inventory models that are currently accessible only through simulation. We believe that this work will serve as a step towards a general learning-based methodology for steady-state analysis in inventory models, with promising opportunities for future research for more complex supply chain systems.

\section{Acknowledgments}
This research was partially supported by the Israel Science Foundation (ISF) Grant No. 1587/25 awarded to E.S.

\bibliographystyle{plainnat} %


\clearpage
\begin{appendices}

\section{Properties of the Generated Test Set}\label{append:scv_rho_testset1}

\begin{figure}[ht]
    \centering
    \begin{subfigure}[b]{0.48\textwidth}
        \centering
\includegraphics[width=\textwidth]{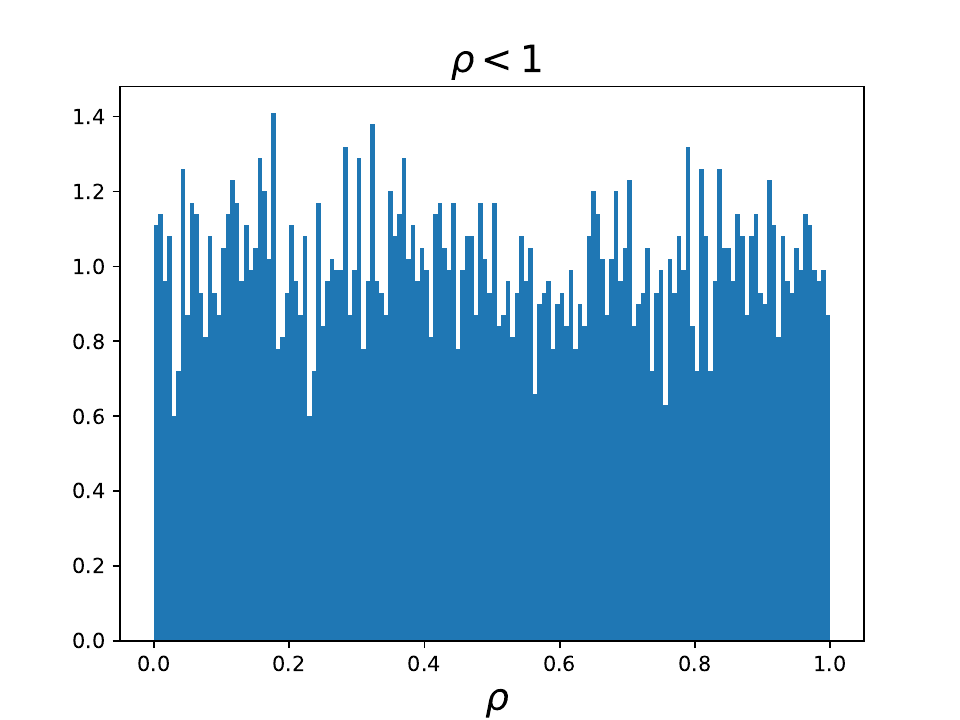}
        \caption{$\rho <1$. }
        \label{fig:rho_small}
    \end{subfigure}
    \hfill
    \begin{subfigure}[b]{0.48\textwidth}
        \centering \includegraphics[width=\textwidth]{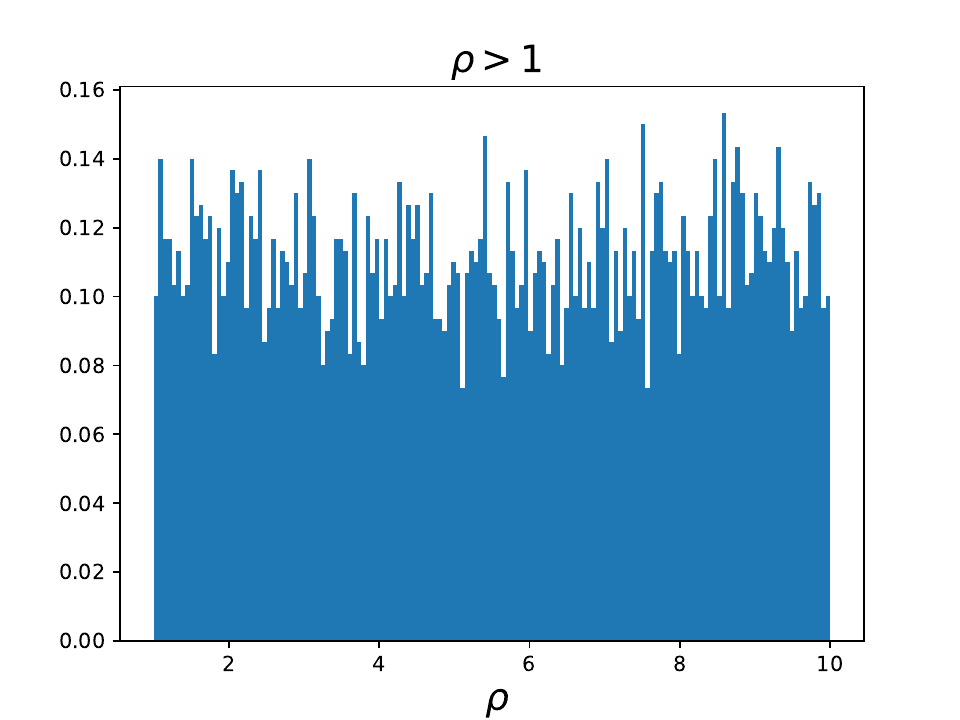}
        \caption{$\rho >1$}
        \label{fig:rho_large}
    \end{subfigure}
    \caption{The ratio $\rho$.}
    \label{fig:rho_diversity}
\end{figure}

In this section, we present the histograms of the statistical metrics of the test sets. Figure ~\ref{fig:rho_diversity} presents the histogram of $\rho$, for $\rho <1$ and $\rho >1$.  We observe a wide range of $\rho$ values; the ratio between demand and delivery time can be lower than one and higher than one, and can take values up to 10.

\begin{figure}[!htp]
\centering
\includegraphics[scale=0.5]{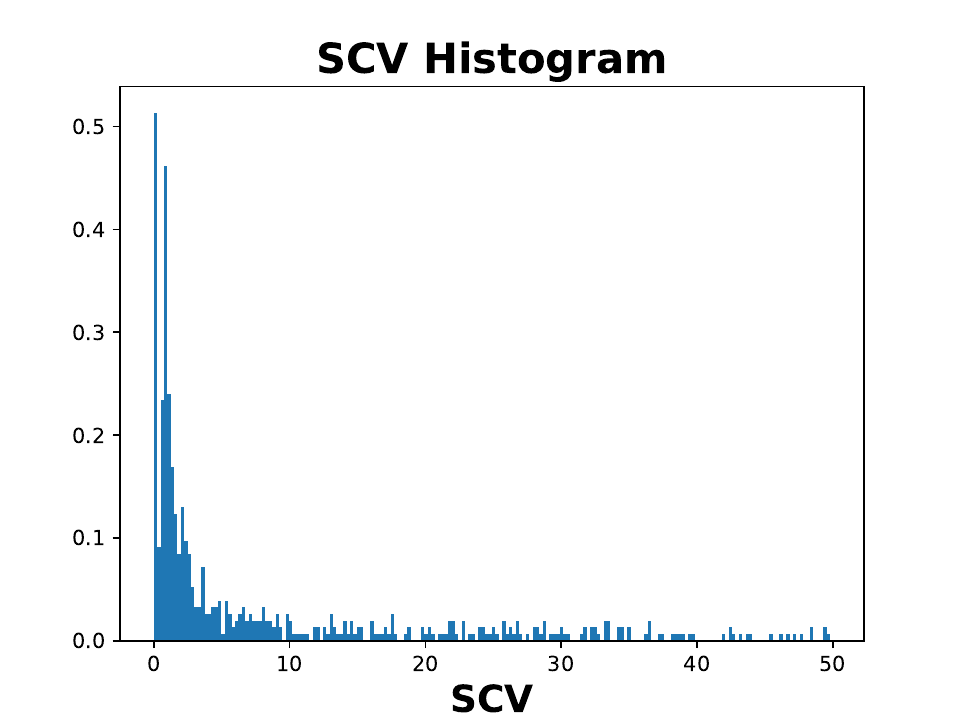}
\caption{The SCV values of the test set. }
\label{fig:scv_hist}
\end{figure}

\begin{figure}[!htp]
\centering
\includegraphics[scale=0.5]{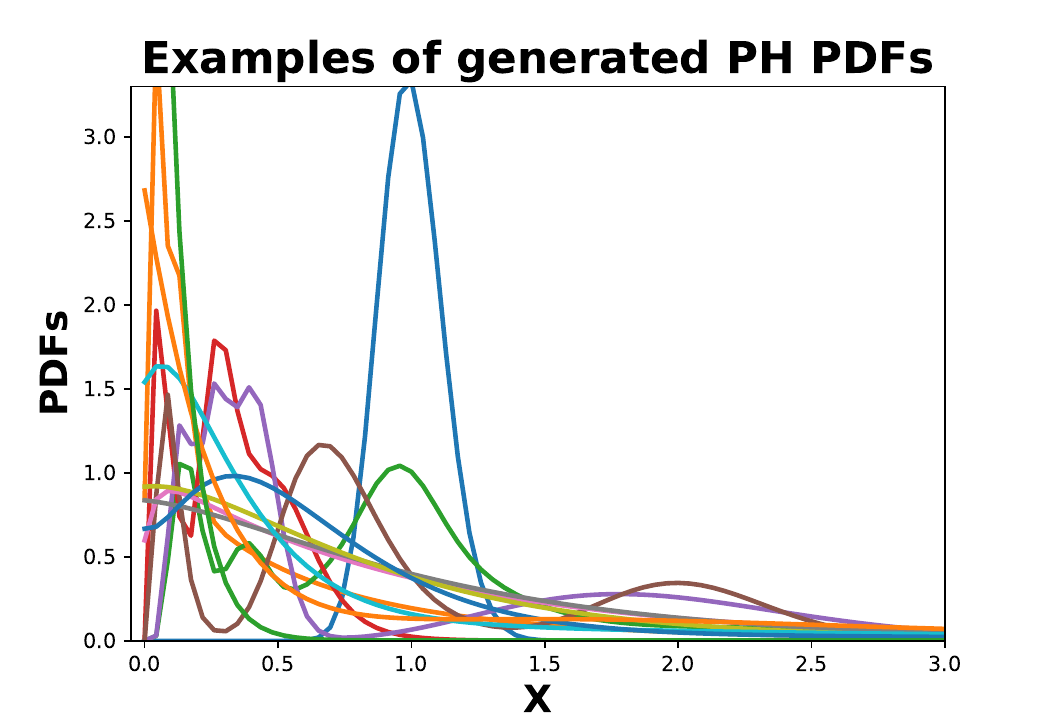}
\caption{Different PDFs of generated PH. }
\label{fig:pdfs}
\end{figure}

Figure~\ref{fig:scv_hist} illustrates the SCV histogram of the generated PH distributions representing  $D$ and $L$ as described in Section~\ref{sec:input_generation}. Figure~\ref{fig:pdfs} presents the probability density functions (PDFs) of 15 examples drawn from the test set. We see that the data set spanned a wide range of distribution shapes.

\section{Training Process}\label{append:train_proc}

Using the SAE metric, we conducted a sensitivity analysis to understand the impact of key hyperparameters on the predictive performance of the NN networks; the analysis was performed separately for \textbf{NN 1}, \textbf{NN 2}, and \textbf{NN 3}. Given their structural similarity, all three NNs yielded similar optimal hyperparameters.

The hyperparameters explored include learning rate, number of training epochs, number of hidden layers, number of neurons per layer, batch size, and the weight-decay parameter used in the Adam optimizer~\cite{8624183}. The search space is defined as follows:
\begin{itemize}
    \item Number of moments:  $\{2,3,4,5,6,7,8,9,10\}$
    \item Learning rate (Lr): $\{0.01, 0.05, 0.001, 0.0001\}$
    \item Number of hidden layers: $[4, 6]$
    \item Neurons per layer: $\{10, 50, 60, 80, 90, 100\}$
    \item Batch size (B): $\{64, 128, 256\}$
    \item Weight decay (Wd): $\{10^{-4}, 10^{-5}, 10^{-6}\}$
\end{itemize}

We performed 200 random searches over the above hyperparameter combinations for each pair of input moment orders, selecting the configuration with the lowest SAE on the validation set. Each run took approximately 150.21 minutes on an NVIDIA GeForce 4070 GPU with 8GB of memory. The optimal configuration for all NNs is given in Table~\ref{tab:nn_config}:

\begin{table}[!htp]\centering
\caption{Congifuration of the NNs}\label{tab:nn_config}
\begin{tabular}{|l|r|r|r|r|r|r|r|} \hline
&\textbf{Num Moments} &\textbf{Lr} &\textbf{Hidden layer} &\textbf{Neurons per layer} &\textbf{B} &\textbf{Wd} \\ \hline
\textbf{NN 1} &5 &0.001 &5 &(50,70,100,90,60) &128 &$10^{-5}$ \\
\textbf{NN 2} &5 &0.001 &5 &(50,70,70,60,50) &128 &$10^{-5}$ \\
\textbf{NN 3} &5 &0.001 &5 &(50,70,70,60,10) &128 &$10^{-5}$ \\ \hline
\end{tabular}
\end{table}

Although the network architecture (depth and width) has some influence, we found that the most significant factors determining the model's accuracy and convergence are the \textbf{learning rate} and the \textbf{batch size}. Proper tuning of these parameters can improve SAE performance by more than 10\%, especially the stability and the quality of the final approximation. However, once the network is sufficiently expressive, further adjustments to the number of layers or neurons yield only marginal gains. It is also worth noting that the number of moments affects the quality of the \emph{target} distribution rather than the quality of the training dynamics. Hence, its impact is primarily on modeling fidelity, not on the degree of convergence of the network.

To promote computational efficiency, we used an early stopping rule based on loss stagnation: training ends when the relative change in loss over 50 epochs falls below $0.001\%$. Formally, letting $\mathcal{L}_t$ denote the loss at epoch $t$, the training process stops when
\[
\frac{|\mathcal{L}_{t-50} - \mathcal{L}_t|}{\mathcal{L}_{t-50}} < 10^{-5}.
\]
This criterion ensures practical convergence and enhances reproducibility by preventing overfitting.

\end{appendices}

\end{document}